\documentclass[10pt,twocolumn,letterpaper]{article}

\usepackage{cvpr}      % To produce the REVIEW version
\usepackage{times}
\usepackage{epsfig}
\usepackage{graphicx}
\usepackage{amsmath,amssymb,amsfonts}
\usepackage{changepage}
\usepackage{booktabs}
\usepackage{multirow}
\usepackage{makecell}
\usepackage[normalem]{ulem}
\useunder{\uline}{\ul}{}
\usepackage{subcaption}
\usepackage{caption}
\usepackage{pifont}
\usepackage{url}
\usepackage{wrapfig}
\usepackage{tcolorbox}
\usepackage{enumitem}
\usepackage{algorithm,algorithmicx,algpseudocode}
\usepackage{color}
\usepackage{booktabs}

\definecolor{cvprblue}{rgb}{0.21,0.49,0.74}
\usepackage[pagebackref,breaklinks,colorlinks,allcolors=cvprblue]{hyperref}

\def\confName{CVPR}
\def\confYear{2026}

\title{Adapting Large VLMs with Iterative and Manual Instructions for
\\
Generative Low-light Enhancement}

\author{Xiaoran Sun\textsuperscript{1,\textdagger}, Liyan Wang\textsuperscript{1,\textdagger}, Yeying Jin\textsuperscript{2, \textdaggerdbl}, Kin-man Lam\textsuperscript{3}, Zhixun Su\textsuperscript{1,\textasteriskcentered}, Yang Yang\textsuperscript{4}
\\
Jinshan Pan\textsuperscript{5}, Cong Wang\textsuperscript{4,\textdaggerdbl}
\\
$^{1}$Dalian University of Technology $^{2}$National University of Singapore 
\\
$^{3}$Hong Kong Polytechnic University $^{4}$University of California, San Francisco
\\
$^{5}$Nanjing University of Science and Technology \\
\{sunxiaoran, wangliyan\}@mail.dlut.edu.cn; jinyeying@u.nus.edu; kin.man.lam@polyu.edu.hk 
\\
yang.yang4@ucsf.edu; zxsu@dlut.edu.cn; \{sdluran, supercong94\}@gmail.com
}



\begin{document}
\maketitle

\footnotetext{\textdagger\ Equal contributions.}
\footnotetext{\textdaggerdbl\ Project leader.}
\footnotetext{\textasteriskcentered\ Corresponding author.}

\begin{abstract}
Most existing low-light image enhancement (LLIE) methods rely on pre-trained model priors, low-light inputs, or both, while neglecting the semantic guidance available from normal-light images. This limitation hinders their effectiveness in complex lighting conditions. In this paper, we propose VLM-IMI, a framework that adapts large vision-language models with iterative and manual instructions for generative LLIE. VLM-IMI mainly contains two branches: Normal-Light Instruction Prior Generation (NL-IPG) and Instruction-aware Light Enhancement Diffusion (IA-LED). The NL-IPG incorporates textual descriptions of the desired normal-light content as enhancement cues, enabling semantically informed restoration. IA-LED incorporates instruction priors from the NL-IPG to guide the diffusion process, enabling precise illumination enhancement. To effectively integrate cross-modal priors, we introduce a learnable instruction prior fusion module, which dynamically aligns and fuses image and text features, promoting the generation of detailed and semantically coherent outputs. During inference, as the ground-truth normal-light images are not available, we propose an inference with an iterative instructions strategy to refine textual instructions, progressively improving visual quality. Our VLM-IMI also inherently supports manual instruction control by allowing users to directly input custom instructions into the LLM to generate user-expected outputs. Experiments across diverse scenarios demonstrate that VLM-IMI outperforms SOTA methods in terms of perception and realism. The source code is available at: \url{https://github.com/sunxiaoran01/VLM-IMI}.
\end{abstract}

\vspace{-6.5mm}
\section{Introduction}\label{sec:intro}
\vspace{-1.5mm}
Low-light image enhancement (LLIE) aims to restore normal-light, noise-free images from dark and noisy inputs, supporting a wide range of downstream tasks~\cite{deng2019arcface,lin2017feature, xu2021exploring}.
Traditional approaches~\cite{land1977retinex,bovik2010handbook,lee2013contrast, li2018structure, cai2023brain} typically enhance brightness and contrast using hand-crafted priors. However, these methods often overlook noise degradation, leading to poor detail preservation and color distortion.
To overcome these limitations, recent research has shifted toward deep learning-based LLIE methods and related fields~\cite{wang2020joint,guo2020zero,lore2017llnet,yi2023diff, wang2022local, wu2022uretinex,jin2022unsupervised,yang2022rethinking,ma2023bilevel,cai2023retinexformer,jin2023enhancing,wang2024correlation,msgnn_ijcai24,PercepLIE_mm25,wang2025ultra,uhd_survey,aaai25_pptformer,ijcv26_uhdpromer}. Despite their success, many of these models learn a direct end-to-end mapping from low-light to normal-light images without auxiliary guidance, which can result in suboptimal visual quality, especially in challenging lighting conditions.
Recently, denoising diffusion probabilistic models (DDPMs)~\cite{ho2020denoising} have emerged as powerful generative frameworks, achieving impressive results in realistic detail synthesis~\cite{hou2023global, jiang2023low, yi2023diff, fei2023generative, lugmayr2022repaint, wang2024zero, wu2024seesr}. These methods typically rely on model priors or low-light inputs~\cite{fei2023generative, wang2024zero} to generate a single output.
%, as illustrated in Figure~\ref{fig:Model comparisons}(a), which summarizes representative diffusion-based approaches.
%
However, such architectures do not take the diverse semantics and illumination conditions of real-world normal-light scenes into model development. As a result, they may produce under- or over-exposed outputs when applied to varying lighting scenarios.
% , as shown in Figures~\ref{fig:Visual comparisons}(b)–(f).
%

% Natural language has been applied to vision to boost performance by using large Vision-Language Models (VLMs)~\cite{alayrac2022flamingo, li2022blip, liu2023visual, radford2021learning}.
% %
% For image enhancement, some research has demonstrated that the language can improve the enhanced quality, which provides additional priors to models~\cite{luo2023controlling, wu2024seesr, li2024light}.
% %
% However, current approaches do not examine the dynamic results when facing different inputs, especially for low-light images with different illumination.
% %
% Hence, \textit{can we process images with different lighting and produce different results depending on the lighting by adapting language instructions?} 
Natural language has recently been integrated into vision tasks through large vision-language models (VLMs)~\cite{alayrac2022flamingo, li2022blip, liu2023visual, radford2021learning}, significantly enhancing performance by providing rich semantic priors. In the context of image enhancement, emerging studies have shown that language-based guidance can improve visual quality by introducing additional contextual cues~\cite{luo2023controlling, wu2024seesr, li2024light}.
However, existing approaches often overlook the dynamic nature of visual inputs, particularly in low-light scenarios with varying illumination conditions. These methods typically either produce static outputs, failing to adapt enhancement strategies based on input-specific lighting characteristics, or inadequately exploit cues from normal-light images to guide the enhancement process.
To address these challenges, we propose VLM-IMI, a framework that adapts a large vision-language model with iterative and manual instructions for low-light image enhancement.
%
%VLM-IMI comprises two key components: Instruction Prior Parsing (I2P) branch and Instruction-aware Lighting Diffusion (ILD) branch. 
VLM-IMI comprises two key components: Normal-Light Instruction Prior Generation (NL-IPG) branch and Instruction-aware Light Enhancement Diffusion (IA-LED) branch.
The NL-IPG branch extracts and encodes textual instructions that describe lighting characteristics from normal-light images using a pre-trained vision-language model and a large language model (LLM) encoder. These instructions serve as semantic priors to guide the enhancement process.
The IA-LED branch integrates the extracted instruction priors with normal-light images to guide the diffusion process, enabling precise illumination enhancement while preserving structural integrity and semantic consistency.
To effectively integrate cross-modal information, we introduce a learnable instruction prior fusion module that dynamically aligns and fuses image and text features. This facilitates the generation of visually detailed and semantically coherent outputs.
%
% Furthermore, during the inference phase, we introduce an iterative feedback strategy to continuously refine the text prompts corresponding to the enhanced images. This strategy improves the precision and contextual relevance of the prompts, even in the absence of ground-truth images, ultimately contributing to the generation of high-quality visual outputs.

% Furthermore, we introduce an iterative instruction strategy to continuously refine the text instructions corresponding to the enhanced images at the inference phase. 
% %
% This strategy improves the precision and contextual relevance of the instructions, ultimately contributing to the generation of high-quality visual outputs.
% %
% Additionally, our method allows for precise control over the diffusion denoising process by incorporating manually provided text information, enabling controlled image enhancement without relying on a visual-language model and generating results with human instructions solely using an LLM.
% %
% Figure~\ref{fig:Visual comparisons} shows our method effectively enhances both global and local contrast, prevents overexposure in well-lit regions, and avoids color distortions or artifacts, particularly when faced with diverse low-light scenarios, thereby improving the model's robustness and generalizability. 
However, ground-truth is not available for getting information about normal-light images at inference. To solve this challenge, we propose an iterative instruction strategy to progressively refine textual guidance based on the enhanced outputs. This refinement improves the precision and contextual alignment of the instructions, thereby contributing to higher-quality visual results.
In addition, our framework supports manual instruction control during inference, allowing users to guide the diffusion process using purely textual instructions. This enables flexible enhancement relying solely on LLMs for instruction-driven generation rather than relying on vision-language models.
%
% As illustrated in Figure~\ref{fig:Visual comparisons}(g)-(i), our method effectively enhances both global and local contrast, mitigates overexposure in well-lit regions, and avoids color distortions or artifacts, and also produces various outputs guided by different instructions.
%
% As illustrated in Figure~\ref{fig:Visual comparisons}(f)-(h), our method effectively enhances both global and local contrast, mitigates overexposure in well-lit regions, and avoids color distortions or artifacts. These capabilities demonstrate the robustness and generalizability of our approach across diverse low-light conditions.
%
% Extensive experimental results show that, compared to state-of-the-art methods, our approach achieves excellent quantitative and qualitative performance on different datasets. 

Our contributions can be summarized as follows:
\begin{itemize}
\item 
%We propose PromptDiff, a novel multi-modal prompt learning method based on diffusion that leverages the generative prior of Stable Diffusion to enhance the generalization, adaptability, and realism of low-light image enhancement.
We propose VLM-IMI, a large VLM framework with iterative and manual instructions, incorporating normal-light language instructions into the diffusion model to generate various images with different instructions to effectively handle complex low-light scenarios.
%\item 
%We explore the significance of semantic priors in diffusion-based LLIE by designing instruction templates to guide the vision-language model in generating appropriate text prompts, or allowing users to customize text prompts, which are then refined through an iterative strategy, thus providing the model with flexible and convenient text control capabilities.
% \item 
%We propose a semantic priors guider to extract semantic priors from a pre-trained large language model and introduce temporal information by integrating time steps in the normalized adaptive layer, dynamically interacting with text features to better guide the noise prediction at each diffusion stage.
% We design a novel Semantic Prior Fusion Module (SPFM) that dynamically interacts with text features through learnable prompts and incorporates timesteps within a normalized adaptive layer, effectively learning contextual semantic information.
\item We propose an iterative instruction strategy during inference to continuously refine text instructions associated with enhanced images. Furthermore, our model allows users to manually customize the instructions, thereby achieving controlled enhancement effects.

\item Experiments conducted on several well-known datasets demonstrate that our proposed VLM-IMI surpasses existing state-of-the-art LLIE methods in terms of naturalness and realism, especially in real-world low-light scenarios.
\end{itemize}
\begin{figure*}[!t]
% \vspace{-2mm}
    \centering
    \includegraphics[width=\textwidth]{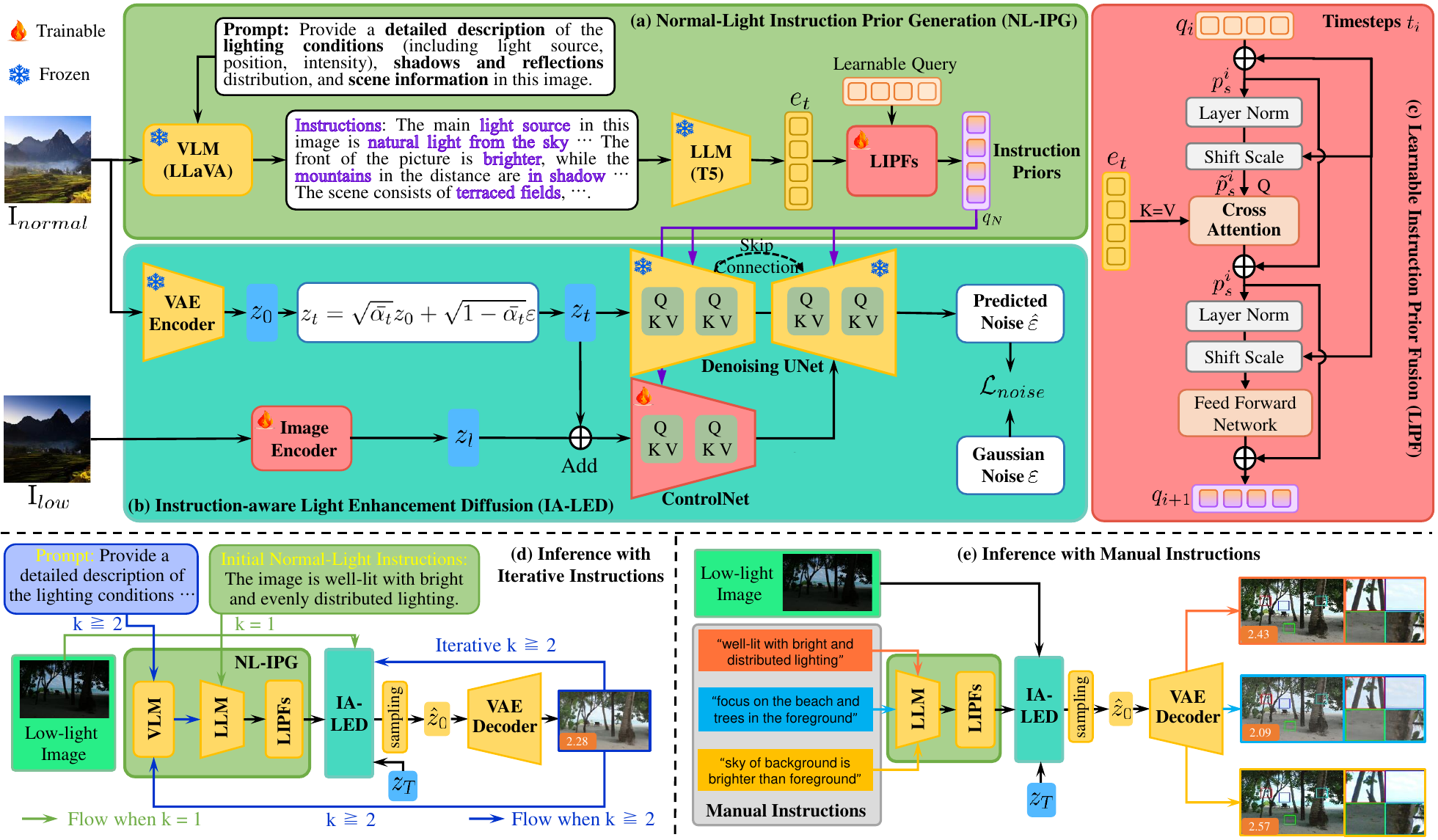}
   % \put(-327,195.5){\tiny{$\mathcal{P}$}}
    %\put(-300,83){\tiny{$\mathit{z_{l}}$}}
     \vspace{-6.5mm}
    \caption{\textbf{Overall framework of VLM-IMI}, which mainly contains (a) Normal-Light Instruction Prior Generation (NL-IPG) and (b) Instruction-aware Light Enhancement Diffusion (IA-LED). In the NL-IPG, we employ the VLM to generate text instructions containing lighting conditions, scene content, and other information from normal-light images. Text features are derived from input into the LLM text encoder, and the (c) Learnable Instruction Prior Fusion (LIPF) is used to effectively integrate text embeddings. In the IA-LED, the low-light images and the above text embeddings serve as conditional inputs to the controlled diffusion model.
    % , which follows the defined workflow. 
    During inference, we propose an iterative instruction strategy (d) to continuously refine text instructions associated with the enhanced images. Furthermore, our model also allows users to manually customize the instructions, thereby achieving controlled enhancement effects (e).
    % A large vision-language model (VLM) extracts textual description instructions from normal-light images, capturing lighting conditions, scene semantics, and contextual cues. These instructions are then encoded using a large language model (LLM) text encoder.
    % %
    % Instruction Prior Fusion Module (IPFM) facilitates cross-modal interaction between textual and visual features, effectively integrating the resulting text embeddings.
    %To effectively integrate the resulting text embeddings, we employ the Instruction Prior Fusion Module (see (b)), which facilitates cross-modal interaction between textual and visual features.
%
% (b) Instruction Prior Fusion Module. This module dynamically aligns and fuses the encoded text features with image representations, enabling the model to generate semantically coherent and visually detailed outputs.
%
%(c) Inference Pipeline with Iterative and Manual Instructions: during inference, we introduce an iterative instruction strategy that continuously refines textual guidance based on progressively enhanced images. Additionally, the model supports manual instruction customization, allowing users to directly influence the enhancement process through natural language instructions.
% , thereby achieving controllable and user-specific enhancement effects.
    }
    \label{fig:pipeline}
    \vspace{-3mm}
\end{figure*}

%\vspace{-2mm}
\section{Related Works}\label{sec:formatting}
%\vspace{-1.5mm}
{\flushleft \bf Low-light image enhancement.}
Numerous works have been proposed to improve the visual quality and perceptibility of poorly illuminated images. 
Traditional methods rely on histogram equalization~\cite{bovik2010handbook} and Retinex theory~\cite{land1977retinex}. 
% HE-based methods \cite{lee2013contrast} enhance low-light images by changing the histogram of the images. Retinex-based approaches \cite{li2018structure, cai2023brain} decompose an image into a reflectance map and an illumination map, then either use the reflectance map as the final output or adjust the dynamic range of the illumination map to improve visual quality. 
However, when applied to complex real-world images, these methods often introduce local color distortions~\cite{wang2019underexposed}.
Recently, learning-based LLIE methods \cite{wang2022local, guo2020zero, zhang2021beyond, yi2023diff, wu2022uretinex, ma2022toward} have achieved remarkable results, demonstrating greater robustness compared to traditional methods. 
These methods usually adopt an end-to-end learning framework to directly map low-light images to normal-light ones. 
However, they often fail to handle images with varying illumination, leading to underexposed or overexposed results.
% For example, Lore \etal \cite{lore2017llnet} proposed LLNet, which utilizes a stacked sparse denoising auto-encoder for LLIE. 
% Lv \etal \cite{lv2018mbllen} presented a CNN-based approach that extracts features at different levels through a multi-branch network and produces the final enhanced result via multi-branch fusion. Wei \etal \cite{wei2018deep} combined Retinex theory with deep learning methods, designed a Retinex-Net to perform the reflectance/illumination decomposition and low-light enhancement jointly. Yang \etal \cite{yang2020fidelity} proposed a semi-supervised method to obtain a linear band representation of an enhanced image. EnlightenGAN \cite{jiang2021enlightengan} is the first unsupervised work for LLIE based on generative adversarial network. 
% \begin{figure}[!t]
%     \centering
%     \includegraphics[width=0.5\textwidth]{LaTeX/images/Figure2-2.pdf}
%     \caption{\textbf{Inference Pipeline with Iterative and Manual Instructions}. During inference, we introduce an iterative instruction strategy that continuously refines textual guidance based on progressively enhanced images. Additionally, the model supports manual instruction customization, allowing users to directly influence the enhancement process through natural language instructions.} \label{fig:sample_process}
%      %\vspace{-3mm}
% \end{figure}

%-------------------------------------------------------------------------
% \subsection{Diffusion-based Image Restoration}
{\flushleft \bf Diffusion-based image restoration.}
Diffusion models have recently gained attention in the field of low-level vision~\cite{li2024light,ho2020denoising,ozdenizci2023restoring,rombach2022high,yi2023diff}, due to their ability to generate high-quality outputs through an iterative denoising process. These models have shown great potential in image restoration tasks, such as inpainting \cite{lugmayr2022repaint}, deblurring \cite{whang2022deblurring}, super-resolution \cite{qu2024xpsr,wu2024seesr}, JPEG compression \cite{guo2025compressionawareonestepdiffusionmodel}, and low-light image enhancement \cite{hou2023global,jiang2023low,wang2024zero}. 
Additionally, some methods~\cite{liu2023accelerating,liu2024diff,ruiz2023dreambooth} introduce conditional information (\eg, labels, text, or other modal inputs) to guide diffusion models using additional priors for better restoration. 
However, these methods usually fail to generate visually pleasing results in images with varying illumination, and they lack the flexibility needed to handle complex low-light scenarios.
% To make the generated images more precise and controllable, ControlNet \cite{zhang2023adding} adds spatial conditional control to large pre-trained text-to-image diffusion models, generating high-quality images and solving the problem of uncontrollable output.
%-------------------------------------------------------------------------
% \subsection{Large Language Model in Vision}
{\flushleft \bf Large language models in restoration.}
With advancements in vision-language models (VLMs)~\cite{alayrac2022flamingo, li2022blip, liu2023visual, radford2021learning} and large language models (LLMs)~\cite{lee2018pre,raffel2020exploring,touvron2023llama,liu2023visual}, many image restoration works~\cite{luo2023controlling, wu2024seesr, li2024light} have integrated VLMs to understand image content and provide rich cross-modal priors to guide the restoration process. 
%
% CLIP~\cite{radford2021learning} leverages contrastive learning to align visual features and textual features, while LLaVA~\cite{liu2023visual} demonstrates impressive capabilities in multi-modal visual question answering. Thanks to the language understanding capability of large language models (LLMs), \eg, BERT~\cite{lee2018pre}, T5~\cite{raffel2020exploring}, and  Llama~\cite{touvron2023llama}, many text-to-image generation methods~\cite{balaji2022ediff, qu2024xpsr} leverage rich instruction priors encapsulated in LLMs to extract precise text embeddings, which are then converted into guiding signals for image generation.
% Although these excellent works, the exploration of adopting normal-light image into LLIE is still not taken. 
% %
% How to integrate text instructions into restoration for LLIE deserves to study.
Despite these remarkable advances, the integration of normal-light images into LLIE remains largely unexplored. Moreover, the incorporation of textual instructions into LLIE presents a promising yet underexplored research direction.

% \vspace{-1.5mm}
\section{Methodology}
%-------------------------------------------------------------------------
% \vspace{-1.5mm}
% \subsection{Framework Overview}
{\flushleft \bf Framework overview.}
Figure~\ref{fig:pipeline} illustrates the overall pipeline of our VLM-IMI, which contains two key branches: Normal-Light Instruction Prior Generation (NL-IPG) and Instruction-aware Light Enhancement Diffusion (IA-LED). 
Given a paired low-light image $\mathbf{I}_{\mathit{low}}$ and normal-light image $\mathbf{I}_{\mathit{normal}}$, the NL-IPG branch leverages a large VLM, \eg, LLaVA~\cite{liu2023visual}, to generate illumination description instructions based on $\mathbf{I}_{\mathit{normal}}$. 
These instructions are subsequently fed into the LLM (\eg, T5~\cite{raffel2020exploring}) text encoder to obtain text embeddings $e_t$.
In the IA-LED branch, we employ the powerful pre-trained Stable Diffusion model~\cite{rombach2022high} as the backbone, with the extracted instruction priors acting as control signals to guide the conditional enhancement process.
Additionally, we adopt ControlNet~\cite{zhang2023adding} as a controller to ensure the enhanced result retains the same spatial and structural features as $\mathbf{I}_{\mathit{low}}$.
To facilitate the fusion of cross-modal priors between text and image, we introduce a learnable instruction prior fusion module that dynamically interacts with image features and textual ones, encouraging the model to generate detailed and semantically accurate results.

%\vspace{-1mm}
%-------------------------------------------------------------------------
\subsection{Normal-light instruction prior generation} \label{sec:TGB}
%
%\vspace{-1mm}
% The instruction prior parsing (I2P) branch plays a crucial role in the image enhancement process by parsing textual information, which consists of two key components: the instruction generation and the instruction prior fusion module. 
% %
% The instruction generation is used to provide semantic guidance for low-light image enhancement, while the instruction prior fusion module is used to extract and enhance semantic priors to further refine the image restoration process. 
% %
% Together, these components enable the model to generate high-quality results that are consistent with the semantic information provided by the text instructions.
The NL-IPG branch is essential for LLIE, operating by parsing textual information through two main components: an instruction generation module and a learnable instruction prior fusion module. 
The instruction generation module provides semantic guidance for LLIE, while the learnable instruction prior fusion module extracts and enhances these semantic priors to refine the image restoration process. 
Together, these modules enable the generation of high-quality results that align with the textual instructions.

%-------------------------------------------------------------------------
% \subsubsection{Text Prompts Generation}
\begin{figure*}[!t]
% \vspace{-5mm}
    \centering
    \includegraphics[width=\textwidth]{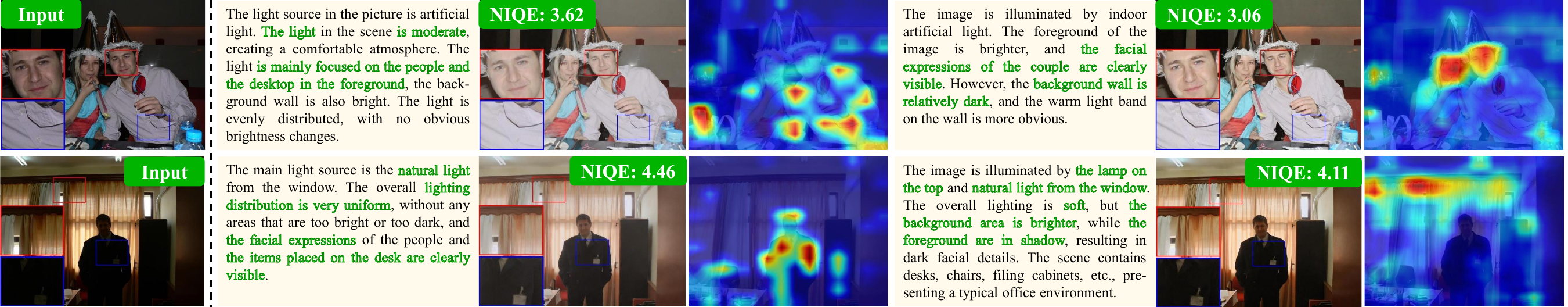}
    \vspace{-7mm}
    \caption{\textbf{Outputs can be controlled by manual instructions}.
    Text instructions can control lighting enhancement and produce results under different lighting conditions. The corresponding Grad-CAM heatmaps~\cite{2017Grad} highlight the model’s attention areas influenced by the text instructions, such as faces or background regions, showing how instructions affect visual enhancement.
    }
    \label{fig:controllability}
     \vspace{-3mm}
\end{figure*}

%%%%%%%%%%%%%%%%%%%%%%%%%%%%%%
%\vspace{-1mm}
\subsubsection{Instruction generation}
%\vspace{-1mm}
% {\flushleft \bf Instruction Generation.} 
% Large VLMs leverage the strengths of both vision and language understanding \cite{alayrac2022flamingo, li2022blip}, enabling them to process cross-modal information and exhibit impressive semantic comprehension capabilities, thanks to large-scale datasets and powerful model architectures. 
% %
% Thus, we employ LLaVA~\cite{liu2023visual} and, by designing appropriate instructions, prompt the VLM to generate textual description instructions of key details, such as light source, illumination intensity, brightness, shadows, and contextual information from normal-light images. 
% %
% These instructions are then encoded using a large language model (LLM) text encoder T5~\cite{raffel2020exploring}.
Large VLMs leverage the strengths of both vision and language understanding \cite{alayrac2022flamingo, li2022blip}. Powered by large-scale datasets and robust model architectures, VLMs can process cross-modal information and exhibit impressive semantic comprehension capabilities. Leveraging this strength, we employ LLaVA~\cite{liu2023visual} and design specific instructions to prompt the VLM. 
%This process generates detailed textual instructions—covering key scene attributes such as light source, illumination intensity, brightness, shadows, and contextual information—from normal-light images. 
This process generates detailed textual instructions from normal-light images, covering key scene attributes such as light source, illumination intensity, brightness, shadows, and contextual information.
Subsequently, these instructions are encoded using the LLM text encoder, T5~\cite{raffel2020exploring}.
%\vspace{-1mm}
{\flushleft \bf Instruction prompt.} Crafting precise instructions for large VLMs requires continuous refinement to align model outputs with human expectations \cite{cheng2023black}. 
To acquire lighting-related instruction priors, we formulate a key prompt for LLaVA~\cite{liu2023visual} as follows: 
\begin{center}
%\vspace{-2.5mm}
\begin{tcolorbox}[colback=blue!10,,%gray background
                  colframe=black,% black frame colour
                  width=8.4cm,% Use 8cm total width,
                  arc=1mm, auto outer arc,
                  boxrule=0.5pt,
                  % title=Instruction Prompt
                 ]
Provide a detailed description of the lighting conditions (including light source, position, intensity), shadows and reflections distribution, and scene information in this image.
\end{tcolorbox}
%\vspace{-2.5mm}
\end{center}
% ``\texttt{Provide a detailed description of the lighting conditions (including light source, position, intensity), shadows and reflections distribution, and scene information in this image}." 
%
%
% \begin{itemize}
% \item  Provide a detailed description of the lighting conditions (including light source, position, intensity), shadows and reflections distribution, and scene information in this image.
% \end{itemize}
% Our experiments show that this instruction consistently yields high-quality outputs from LLaVA, significantly improving enhancement results.
% In practice, we observe that such structured instructions enable LLaVA to generate highly consistent and relevant outputs, effectively capturing the semantic essence of the scene. This iterative process of refining the instructions ensures that the generated priors are rich in detail and highly effective for guiding subsequent image enhancement tasks.
In practice, we observe that these structured instructions significantly enhance LLaVA's output consistency and relevance, enabling it to capture the semantic essence of the scene effectively. This iterative refinement process ensures that the generated priors are rich in detail and highly effective for guiding subsequent image enhancement tasks.

\vspace{-1.5mm}
{\flushleft \bf Function of instruction.} To clarify the function of the defined instruction prompt, we summarise the reasons why we formulate such prompts as follows:
\begin{center}
\vspace{-2.5mm}
\begin{tcolorbox}[%colback=gray!10,%gray background
                    colback=blue!8,
                  colframe=black,% black frame colour
                  width=8.4cm,% Use 8cm total width,
                  arc=1mm, auto outer arc,
                  boxrule=0.5pt,
                  % title=Function of Instruction
                 ]
{\bf Lighting Information.} Lighting information helps the model analyze the light source type (\eg, natural or artificial), location, and intensity (\eg, bright, moderate, soft).
\\
\vspace{-2mm}
\\
{\bf Shadows and Reflections.} The model describes shadow characteristics, including direction, and any visible reflections, etc., aiding in the understanding of light-environment interactions. 
\\
\vspace{-2mm}
\\
{\bf Spatial and Contextual Information.} LLaVA identifies high-level scene information such as scene elements, spatial relationships, and contextual details.
\end{tcolorbox}
\vspace{-2.5mm}
\end{center}
% \begin{itemize}
% \item  { Lighting Information.} Lighting Information helps the model analyze the light source type (\eg, natural or artificial), location, and intensity (\eg, bright, moderate, soft).

% \item  { Shadows and Reflections.} The model describes shadow characteristics, including direction, and any visible reflections, etc., aiding in the understanding of light-environment interactions. 

% \item  { Spatial and Contextual Information.} LLaVA identifies high-level scene information such as scene elements, spatial relationships, and contextual details.
% \end{itemize}
% \begin{itemize}
% \item  
% \item 
% \item 
% \end{itemize}

%-------------------------------------------------------------------------
% \subsubsection{Semantic Priors Guider}

% {\flushleft \bf Instruction Prior Fusion Module.}

\subsubsection{Learnable instruction prior fusion}
%
% Inspired by the Perceiver Resampler~\cite{alayrac2022flamingo}, we propose an Instruction Prior Fusion Module (IPFM), shown in Figure~\ref{fig:pipeline}(b), to effectively incorporate instruction context. 
% %
% This module explores the latent learnable queries to interact with text features through cross-attention layers. 
Inspired by the Perceiver Resampler \cite{alayrac2022flamingo}, we propose a Learnable Instruction Prior Fusion (LIPF) module, as illustrated in Figure \ref{fig:pipeline}(b), specifically designed to effectively integrate the instruction context. 
The LIPF achieves this by using latent, learnable queries to interact with the text features with the following subsequent operations: embedding timestep, integrating semantic context, and enhancing fusion representation.
\vspace{-1.5mm}
{\flushleft \bf Embedding timestep.}
% By integrating timestep information, the module is able to extract dynamic text features, enhancing the model's ability to condition the noise prediction at various stages of the diffusion process.
Integrating timestep information allows the module to encode timestep-aware textual features,
%extract dynamic text features, 
significantly enhancing the model's ability to condition noise prediction at various stages of the diffusion process. 
%
%Specifically, for the $\mathit{i}$-th block, the timestep $t_{\mathit{i}}$ is first embedded and then added with the input $p_s^i$.
Specifically, for the $\mathit{i}$-th block ($i = 1, \cdots, N$), the timestep $t_{\mathit{i}}$ is first embedded and then added with the learnable query vector $q_i$ to obtain a new input $p_s^i$.

% let the input be represented as
%
% Specifically, for the $\mathit{i}$-th block, let the input be represented as: $p_s^i = q_{\mathit{i}}+\mathbf{Embed}(t_{\mathit{i}})$, 
% % \begin{equation}
% %     p_s^i = q_{\mathit{i}}+\mathbf{Embed}(t_{\mathit{i}}),
% % \end{equation}
% where $\mathbf{Embed}(\cdot)$ is a function that encodes the timestep $t_{\mathit{i}}$ as the time embedding.

\vspace{-1.5mm}
{\flushleft \bf Integrating semantic context.}
We first apply an AdaLN layer~\cite{DiT23}, composed of Layer Norm and Shift Scale operations, to dynamically normalize the input $p_s^i$,
% , followed by cross-attention with text embeddings $e_t$.
%
%The output is then combined with the original input via a residual connection
producing an updated representation $\tilde{p}_s^i$.
Next, the $\tilde{p}_s^i$ serves as the Query, and the concatenation of $\tilde{p}_s^i$ and text embeddings $e_t$ serves as the Key and Value to perform fusion by cross-attention with the residual connection by $p_s^i$.
% , formulated as follows:
% \begin{equation}
%     \bar{p}_s^i  = \mathbf{AdaLN}\Big(p_s^i, \mathbf{Embed}(t_i)\Big), 
% \end{equation}
% %
% \begin{equation}
%     \tilde{p}_s^i = p_s^i + \mathbf{CA}\Big(q = \bar{p}_s^i, \, kv = \mathbf{concat}\left[ \bar{p}_s^i, \, e_t\right]\Big),
% \end{equation}
% where $\mathbf{CA}(\mathit{q}, \mathit{kv})$ represents the cross-attention layer. 
%
This mechanism enables the model to integrate semantic context by aligning and refining its representations based on textual instructions, thus adapting its understanding according to the provided instruction priors. 

\vspace{-1.5mm}
{\flushleft \bf Enhancing fusion representation.} The updated representation then proceeds to the next AdaLN layer, followed by a feed-forward network (FFN) to enhance the representation of fused features. 
Finally, a residual connection is applied, and the output is passed to the next block as $q_{i+1}$.
% , formulated as: $q_{i+1} = \tilde{p}_s^i + \mathbf{FFN}\Big(\mathbf{AdaLN}\big(\tilde{p}_s^i, \mathbf{Embed}(t_i)\big)\Big)$.
% \begin{equation}
%     q_{i+1} = \tilde{p}_s^i + \mathbf{FFN}\Big(\mathbf{AdaLN}\big(\tilde{p}_s^i, \mathbf{Embed}(t_i)\big)\Big).
% \end{equation}

% The instruction prior fusion module integrates timestep information and incorporates text features through cross-attention, facilitating semantic conditioning guidance for the diffusion model.

%-------------------------------------------------------------------------
%\vspace{-1mm}
\subsection{Instruction-aware light enhancement diffusion}
%\vspace{-1mm}
%We present the design of the lighting enhancement branch, which uses Stable Diffusion (SD) \cite{rombach2022high} as the backbone network and incorporates the instruction priors provided by the I2P branch.
% We present an instruction-aware lighting diffusion (ILD) branch built upon the Stable Diffusion (SD) model~\cite{rombach2022high}, incorporating instruction priors from the I2P branch. 
% %
% To improve low-light image enhancement, we introduce the ControlNet architecture \cite{zhang2023adding}, which utilizes a trainable copy of the U-Net encoder to learn additional conditional information from low-light images while freezing the backbone diffusion model. 
% %
% Instruction priors are incorporated into the diffusion process via a cross-attention mechanism, providing semantic guidance for enhancement.
We present the IA-LED branch, built upon the Stable Diffusion (SD) model~\cite{rombach2022high}, which incorporates instruction priors from the NL-IPG branch. 
% To effectively enhance low-light images, we integrate the ControlNet architecture \cite{zhang2023adding} that utilizes a trainable copy of the U-Net encoder to learn conditional information directly from the low-light image input. 
\vspace{-1.5mm}
{\flushleft \bf Semantic guidance by instruction priors.}
Crucially, the instruction priors generated from NL-IPG are incorporated into the diffusion process via cross-attention, providing essential semantic guidance for enhancement. Note that the diffusion model remains frozen during training.
\vspace{-1.5mm}
{\flushleft \bf Structural consistency by adopting ControlNet.}
% Beyond instruction priors, structural priors also play a crucial role in reconstructing the normal-light image. To preserve structural consistency, we pass the low-light image through a trainable image encoder, whose architecture mirrors that in \cite{zhang2023adding}, to obtain the latent embedding $\mathit{z_l}$.
% %
% This embedding is then combined with sampled noise $\mathit{z_t}$ to serve as the input to ControlNet. 
% %
% The outputs of ControlNet are subsequently added back into the original U-Net decoder to refine the final enhancement.
% Beyond the instruction priors, structural priors also play a crucial role in reconstructing the normal-light image. To preserve structural consistency, we utilize a trainable image encoder, mirroring the architecture in \cite{zhang2023adding}, to obtain the latent structural embedding $\mathbf{z}_l$ from the low-light image. This embedding $\mathbf{z}_l$ is subsequently combined with the sampled noise $\mathbf{z}_t$ and passed as the input to ControlNet. The resulting outputs of ControlNet are then added to the features of the original U-Net decoder to refine the final enhancement.
Beyond instruction priors, structural priors are essential for accurately reconstructing normal-light images. To maintain structural consistency, we adopt a trainable image encoder, following the architecture in~\cite{wu2024seesr}, to extract a latent structural embedding $z_l$ from the low-light image. This embedding is combined with the sampled noise $z_t$ and fed into ControlNet~\cite{zhang2023adding}, whose outputs are integrated into the U-Net decoder features to refine the final enhancement results.
% , as follows: 
% \begin{equation}
%     z_l=\mathcal{I}(y),
% \end{equation}
% \subsection{Inference with Iterative and Manual Instructions}
% {\flushleft \bf Inference with Iterative and Manual Instructions.}
% {\flushleft \bf Optimizing a few modules.}
%\vspace{-1mm}
\subsection{Optimizing a few modules}
%\vspace{-1mm}
% During the training stage, the normal-light image is encoded into a latent representation $\mathit{z}_0$ using the pre-trained VAE encoder. Noise is gradually added through the diffusion process, leading to the noisy latent representation $\mathit{z_t}$. 
% %
% Given the diffusion step $\mathit{t}$, the noisy latent representation $\mathit{z_t}$, the low-light image latent representation $\mathit{z_l}$, and the instruction prior $\mathit{p_s}$, we train the proposed VLM-IMI network, denoted as  $\epsilon_{\theta}$, to predict the noise added to the noisy latent $\mathit{z_t}$. The objective function for optimization is: $\mathcal {L} = \mathbb{E}_{z_0, t, z_l, p_s, \epsilon \sim \mathcal{N}(0, \textbf{I})}[||\epsilon -\epsilon _{\theta }(z_t, t, z_l, p_s)||^2_2]$.
% %
% To minimize training cost and use the generative prior of SD, we freeze all the parameters of SD during the training process and only update the IPFMs, ControlNet, and the image encoder.
During the training stage, the normal-light image is first encoded into its latent representation $z_0$ using the pre-trained VAE encoder. Noise is then gradually introduced through the diffusion process, resulting in the noisy latent representation $z_t$. We train the proposed VLM-IMI network, denoted as $\epsilon_{\theta}$, to predict the noise $\epsilon$ added to $z_t$. The network takes the following inputs: the timestep $t$, the noisy latent $z_t$, the low-light image latent $z_l$, and the instruction prior $q_N$. The objective function for optimization is the standard mean-squared error (MSE) loss: $\mathcal {L}_{noise} = \mathbb{E}_{z_0, t, z_l, q_N, \epsilon \sim \mathcal{N}(0, \mathbf{I})}[||\epsilon - \epsilon _{\theta }(z_t, t, z_l, q_N)||^2_2]$.
To minimize training costs and leverage the powerful generative prior of Stable Diffusion, we freeze all parameters of the backbone SD model during training. We only update the parameters of the Learnable Instruction Prior Fusion (LIPF), ControlNet, and the image encoder.

\begin{table*}[t]
% \vspace{-5mm}
\centering
\caption{\textbf{Quantitative comparisons.}
% on the paired LOL \cite{wei2018deep} and LSRW \cite{hai2023r2rnet} datasets, and unpaired DICM \cite{lee2013contrast}, NPE \cite{wang2013naturalness}, and VV \cite{vonikakis2018evaluation} datasets. 
The best and second-best results are highlighted in \textbf{bold} and \underline{underlined}, respectively.
Note $^*$ indicates the model trained on the LOL dataset.}
\vspace{-2mm}
\label{tab:comparison}
\renewcommand{\arraystretch}{1}
% 列格式：删除最左侧的竖线（原为 |c|c|... 改为 c|c|...）
\small % 缩小字体
\setlength{\tabcolsep}{3pt} % 压缩列间距
% 通过缩放适配页面宽度
\resizebox{\textwidth}{!}{ 
\begin{tabular}{c|r|ccc|ccc|cc|cc|cc}
\hline
% 修改所有合并列的左侧竖线（如 |c| 改为 c|）
\multicolumn{2}{c|}{\multirow{2}{*}{\textbf{Datasets}}} &\multicolumn{6}{c|}{\textbf{Synthetic Paired Datasets}}&\multicolumn{6}{c}{\textbf{Real-world Unpaired Datasets}}
\\
\cline{3-14}
\multicolumn{2}{c|}{}& \multicolumn{3}{c|}{LOL \cite{wei2018deep}} & \multicolumn{3}{c|}{LSRW \cite{hai2023r2rnet}} & \multicolumn{2}{c|}{DICM \cite{lee2013contrast}} & \multicolumn{2}{c|}{NPE \cite{wang2013naturalness}} & \multicolumn{2}{c}{VV \cite{vonikakis2018evaluation}} \\ 

\hline
\multicolumn{2}{c|}{Metrics} & PSNR~$\uparrow$ & SSIM~$\uparrow$ & LPIPS~$\downarrow$ & PSNR~$\uparrow$ & SSIM~$\uparrow$ & LPIPS~$\downarrow$ & MUSIQ~$\uparrow$ & NIQE~$\downarrow$ & MUSIQ~$\uparrow$ & NIQE~$\downarrow$ & MUSIQ~$\uparrow$ & NIQE~$\downarrow$ \\ 
\hline
% ========== Traditional Methods ==========
\multirow{14}{*}{\rotatebox[origin=c]{90}{\textbf{Pre-trained Model}}} 
&LIME \cite{guo2016lime} 
& 17.546 & 0.531 & 0.290 & 17.342 & 0.520 & 0.416 & 51.263 & 4.476 & 53.284 & 4.170 & 35.591 & 3.713 \\ 
& CDEF \cite{lei2022low}
& 16.335 & 0.585 & 0.351 & 16.758 & 0.456 & 0.314 & 55.385 & 4.142 & 50.771 & 3.862 & 32.722 & 5.051 \\ 
& BrainRetinex \cite{cai2023brain} 
& 11.063 & 0.475 & 0.327 & 12.506 & 0.390 & 0.374 & 52.774 & 4.350 & 58.116 & 3.707 & 36.289 & 4.031 \\ 
&Zero-DCE \cite{guo2020zero} 
& 14.861 & 0.562 & 0.330 & 15.867 & 0.443 & 0.315 & 60.306 & 3.951 & 62.686 & 3.826 & 39.231 & 5.080 \\ 
& EnlightenGAN \cite{jiang2021enlightengan}
& 17.606 & 0.653 & 0.319 & 17.106 & 0.463 & 0.322 & 59.662 & 3.832 & 62.580 & 3.775 & 30.403 & 3.689 \\ 
& LCDPNet \cite{wang2022local}
& 14.506 & 0.575 & 0.312 & 15.689 & 0.474 & 0.344 & 56.301 & 4.110 & 55.399 & 4.106 & 37.219 & 5.039 \\ 
& SCI \cite{ma2022toward}
& 14.784 & 0.525 & 0.333 & 15.242 & 0.419 & 0.321 & 62.267 & 4.519 & 60.730 & 4.124 & 35.071 & 5.312 \\ 
& BL \cite{ma2023bilevel}
& 10.305 & 0.401 & 0.382 & 12.444 & 0.333 & 0.384 & 55.364 & 5.046 & 57.207 & 4.885 & 37.935 & 5.740 \\ 
%& GDP$^*$ \cite{fei2023generative} & 15.896 & 0.542 & 0.337 & 12.887 & 0.362 & 0.386 & 57.926 & 4.357 & 59.182 & 4.032 & 36.624 & 4.683 \\ 
%\multirow{6}{*}{\rotatebox[origin=c]{90}{\textbf{Trained on LOL}}} % 删除左侧的\multicolumn包裹
%\hline
% ========== Learning-based Methods ==========
%\multirow{21}{*}{\rotatebox[origin=c]{90}{Learning-based}} % 直接使用\multirow
&RetinexNet$^{*}$ \cite{wei2018deep}
& 16.774 & 0.462 & 0.390 & 15.609 & 0.414 & 0.393 & 58.302 & 4.487 & 56.088 & 4.732 & 41.562 & 5.881 \\ 
& DRBN$^{*}$ \cite{yang2020fidelity} 
& 16.677 & 0.730 & 0.252 & 16.734 & 0.507 & 0.376 & 59.558 & 4.369 & 61.933 & 3.921 & 40.174 & 3.671 \\ 
&KinD++$^{*}$ \cite{zhang2021beyond}
& 17.752 & 0.758 & 0.198 & 16.085 & 0.394 & 0.366 & 57.813 & 4.027 & 58.214 & 4.005 & 40.826 & 3.586 \\ 

& URetinexNet$^{*}$ \cite{wu2022uretinex} 
& 19.842 & 0.804 & 0.237 & 18.271 & 0.518 & 0.295 & \underline{64.495} & 4.774 & 63.815 & 4.028 & 40.458 & 3.851 \\ 
& InstructIR$^{*}$ \cite{InstructIR24}
& 22.017	& 0.824	&0.177&	17.709&	0.512&	0.280	&	60.283	&4.755	&61.692&	4.485&	43.603	&4.881\\
& \textbf{VLM-IMI$^{*}$ (Ours)}
& \underline{22.437} & \textbf{0.833 }& 0.172 &18.016  &  \underline{0.537} &\underline{0.188}  &61.791 &3.603  &59.817  &3.518 &40.407  &3.632 \\ 
\hline
% ===== Diffusion-based Subsection =====
\multirow{12}{*}{\rotatebox[origin=c]{90}{\textbf{Trained on Our Dataset}}}
& PairLIE \cite{fu2023learning} 
& 14.853&	0.672&	0.358	 & 13.915	&0.474	&0.317 & 56.636	 & 4.068 & 60.350 & 4.118 &39.910 & 3.824 \\ 
& NeRCo \cite{yang2023implicit}
& 15.279	&0.663&	0.331 & 14.352&	0.503&	0.285 & 57.005&	3.972	&61.113	&3.783	&39.001	&\underline{3.565} \\ 
& CLIP-LIT \cite{liang2023iterative} 
%& 19.039	&0.747&	0.273&	18.701	&0.509	&0.202	&	59.451	&4.120	&62.243	&3.881&	37.826&	3.607
&14.825	&0.524	&0.371	&13.468	&0.411&	0.256	&61.725&	3.889	&63.376	&3.877	&41.768	&3.782\\ 
% & GDP$^*$ \cite{fei2023generative}  % 删除左侧竖线
% & 15.896 & 0.542 & 0.337 & 12.887 & 0.362 & 0.386 & 57.926 & 4.357 & 59.182 & 4.032 & 36.624 & 4.683 \\ 
% & \multicolumn{1}{r|}{} & WeatherDiff$^*$ \cite{ozdenizci2023restoring}
% & 17.913 & 0.771 & 0.272 & 16.507 & 0.487 & 0.431 & 59.210 & 3.773 & 61.371 & 3.677 & 39.852 & \textbf{3.472} \\ 
&Retinexformer \cite{Retinexformer23}&21.184	&0.792	&\underline{0.159}	&16.162	&0.481	&0.267	&60.766&	3.835	&63.846	&3.928	&41.284	&3.875\\
& DiffLL \cite{jiang2023low}
& \textbf{23.214} & \underline{0.825} & 0.212 & 17.617 & 0.485 & 0.249 & 58.268 & \textbf{3.528} & 62.862 & 3.523 & 36.804 & 3.624 \\ 
 & GSAD \cite{hou2023global}
& 17.964	&0.720	&0.205	&	18.526	&0.524&	 0.191&	58.087	&4.445&	60.294	&4.310&	43.808	&4.082 \\ 
 & QuadPrior \cite{wang2024zero}
& 19.571	&0.795	& 0.171&17.874	& 0.536&	0.218&62.716	&4.284	&62.960	&4.247&	\underline{44.895}&	3.967 \\ 
 & Diff-Plugin \cite{diffplugin24}&19.836	&0.684	&0.292&18.542	&0.526&	0.276	&63.150	&3.827	&60.252	&3.575	&42.537	&3.672
\\
 & UniProcessor \cite{UniProcessor24}&20.924	&0.811	&0.187&18.330	&0.534&	0.241	&62.874	&3.643	&62.593	&3.528	&44.390	&3.841\\
 &GPP-LLIE \cite{GPP-LLIE25}
&19.583	&0.769	&0.263	&17.904	&0.508	&0.256	&58.253	&3.597	&60.368	&3.570 &	41.841	&3.772
\\
 &GEFU \cite{GEFU25}&18.648&0.792	&0.197	& \underline{18.961}&	0.536	&0.198	&60.792	&3.622	& \underline{63.893}	&\underline{3.495}	&38.479	&4.064
\\
 & \textbf{VLM-IMI (Ours)}
& 21.112 & 0.802 & \textbf{0.155} & \textbf{19.351} & \textbf{0.548} & \textbf{0.173} & \textbf{64.661} & \underline{3.551} & \textbf{64.097} & \textbf{3.421} & \textbf{45.984} & \textbf{3.523} \\ 
\hline
\end{tabular}
}
%\vspace{-1mm}
\end{table*}
%%%%%%%%%%%%%%%%%%%%%%%%%%%%
%\vspace{-1mm}
\subsection{Inference with iterative and manual instructions}
\vspace{-1.5mm}
{\flushleft \bf Inference with iterative instructions.}
% During the inference stage, where ground-truth (GT) normal-light references are typically unavailable, we propose an iterative instruction refinement strategy (Figure~\ref{fig:pipeline}(c)). 
% %
% This looped process, performed with a fixed network, aims to progressively improve both the accuracy of textual instructions and the quality of lighting enhancement.
% %
% Initially, a lighting-related description $p_s$ is generated by LLM from an initial text $P$. This instruction is used to guide the enhancement process, producing an initial output. The enhanced result $\hat{x}$ is then reintroduced into LLaVA to generate updated textual descriptions reflecting the current lighting and scene context. These refined instructions are fed back into the model, further improving the enhancement quality. 
During the inference stage, where ground-truth (GT) normal-light references are typically unavailable, we employ an iterative instruction refinement strategy. This looped process, executed with a fixed network, is designed to progressively enhance both the accuracy of the textual instructions and the quality of the lighting enhancement. The process is initiated by a designed initial normal-light instruction as follows:
\begin{center}
%\vspace{-2mm}
\begin{tcolorbox}[colback=blue!10,,%gray background
                  colframe=black,% black frame colour
                  width=8.4cm,% Use 8cm total width,
                  arc=1mm, auto outer arc,
                  boxrule=0.5pt,
                  % title=Instruction Prompt
                 ]
The image is well-lit with bright and evenly distributed lighting.
\end{tcolorbox}
%\vspace{-2mm}
\end{center}
This instruction is fed to LLM and used to guide the enhancement, yielding an initial enhanced output. Crucially, this enhanced result is then reintroduced into LLaVA with the designed prompt to generate updated textual instruction descriptions that reflect the current scene's lighting and context. These refined instructions are subsequently fed back into the model, driving further improvements in enhancement quality. Figure \ref{fig:pipeline}(d) illustrates the process of the VLM-IMI's inference with iterative instructions.
% The iterative instruction generation process is as follows:
% \begin{equation}
% % p_s=\mathcal{S}(\mathcal{F}_{\mathrm{L}}(P)) \quad if \quad k = 1, \quad else \quad p_s=\mathcal{S}(\mathcal{F}_{\mathrm{L}}(\mathcal{F}_{\mathrm{V}}(\hat{x})))
%     p_s=\left\{
%     \begin{aligned}
%     \mathcal{S}(\mathcal{F}_{\mathrm{L}}(P)), \quad k = 1\\ 
%     \mathcal{S}(\mathcal{F}_{\mathrm{L}}(\mathcal{F}_{\mathrm{V}}(\hat{x}))), \quad k = 2\\
%     \end{aligned}
%     \right
%     .
% \end{equation}
% where $\mathcal{F}_{\mathrm{V}}$, $\mathcal{F}_{\mathrm{L}}$, and $\mathcal{S}$ denote  pre-trained VLM, LLM, and IPFMs, respectively.
%The full procedure is detailed in Algorithm~\ref{alg:sampling}.

\vspace{-1.5mm}
{\flushleft \bf Inference with manual instructions.}
% In addition to iterative refinement, our method also supports manual instruction control by directly inputting user-defined instructions into the LLM. 
% %
% This enables precise and interpretable control over the diffusion denoising process, allowing for customizable enhancement tailored to diverse low-light scenarios.
% %
% Figure~\ref{fig:controllability} shows that our method enables fine-grained control over enhancement outcomes. 
% %
% The corresponding Grad-CAM heatmaps~\cite{2017Grad} highlight the model’s attention regions—such as faces or background areas—demonstrating that textual instructions effectively influence the enhancement process under varying illumination conditions.
In addition to iterative refinement, our method inherently supports manual instruction control by allowing users to directly input custom instructions into the LLM. This capability offers precise and interpretable control over the diffusion denoising process, enabling customizable enhancement tailored to diverse low-light scenarios. As demonstrated in Figure \ref{fig:controllability}, our method enables fine-grained control over enhancement outcomes. Furthermore, the corresponding Grad-CAM heatmaps \cite{2017Grad} visually confirm the model's selective attention—highlighting regions such as faces or background areas—thereby proving that textual instructions effectively modulate the enhancement process under varying illumination conditions. Figure~\ref{fig:pipeline}(e) shows the process of the VLM-IMI's inference with manual instructions.
% More results are shown in the Appendix.

%\vspace{-1.5mm}
\section{Experiments}\label{sec:Experiments}
%\vspace{-1.5mm}
% In this section, we conduct extensive experiments to examine the effectiveness of our proposed VLM-IMI and deeply analyze the effect of each proposed component of VLM-IMI.
\begin{figure*}[t]
% \vspace{-2mm}
    \centering
    \includegraphics[width=\textwidth]{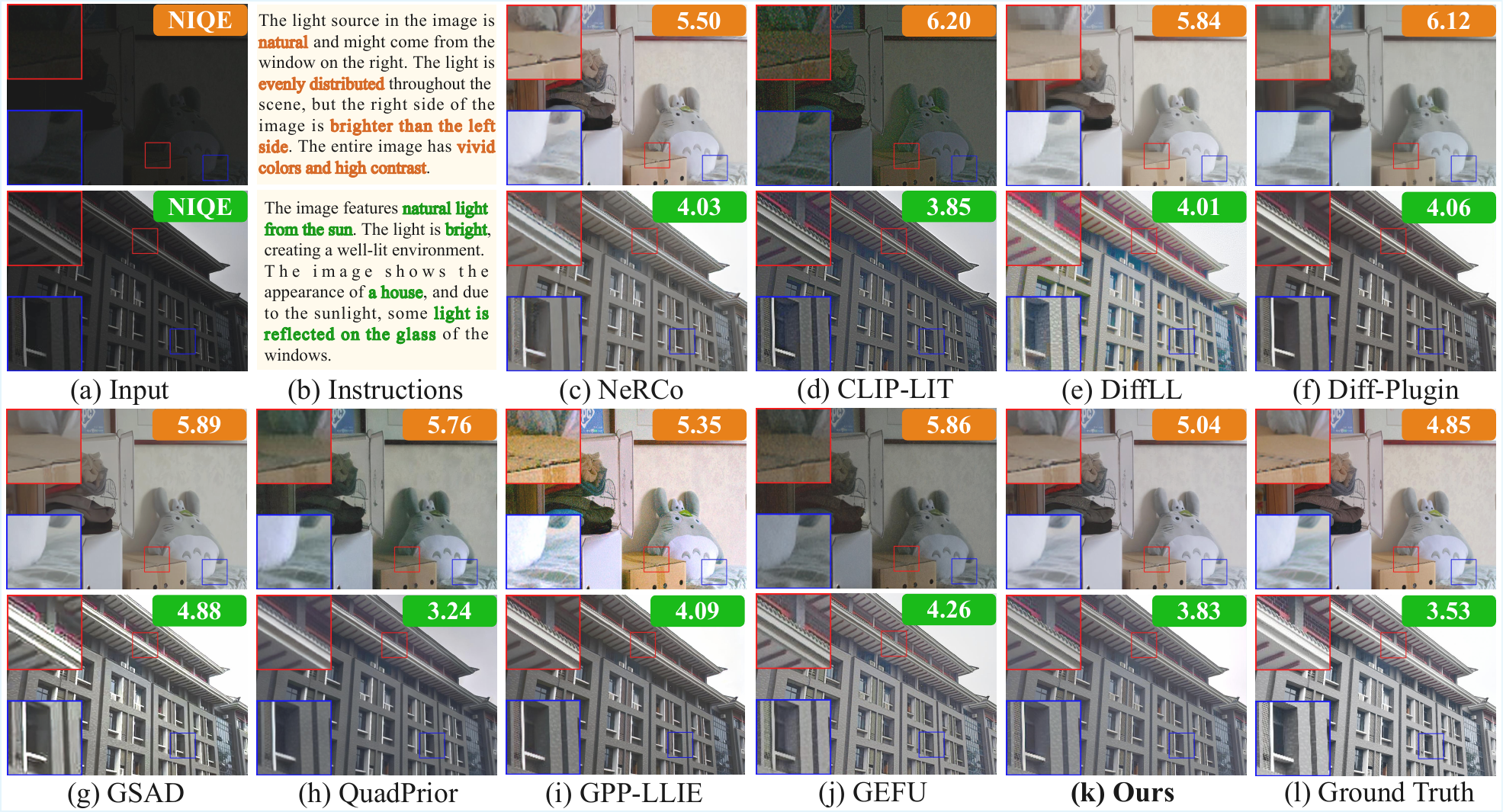}
    \vspace{-5mm}
    \caption{\textbf{Visual comparison} on LOL~\cite{wei2018deep} and LSRW~\cite{hai2023r2rnet}.
    Our VLM-IMI is able to produce more realistic results with sharper structures.
    }
   \label{fig:comparison_paired}
   \vspace{-6.5mm}
\end{figure*}

%%%%%%%%%%%%%%%%%%%%%%%%%%%%%%%%%

\subsection{Experimental setup} \label{sec:settings}

% \noindent{\bf Datasets.} 
\vspace{-1.5mm}
{\flushleft \bf Datasets.}
We train VLM-IMI on a dataset of approximately 15k normal/low-light images collected from RAISE~\cite{dang2015raise}, LSRW~\cite{hai2023r2rnet},  LOLI-Street~\cite{li2024light}, and LOL~\cite{wei2018deep}. To evaluate the performance of our proposed method, we conduct experiments on two paired test datasets, LOL \cite{wei2018deep} and LSRW \cite{hai2023r2rnet}, and three unpaired datasets containing only real low-light images, DICM \cite{lee2013contrast}, NPE \cite{wang2013naturalness}, and VV \cite{vonikakis2018evaluation}.

\vspace{-1.5mm}
{\flushleft \bf Metrics.}
% \noindent{\bf Metrics.} 
As our method aims to enhance realism and perceptual quality, we adopt two non-reference metrics for evaluating naturalness, including MUSIQ~\cite{ke2021musiq} and NIQE~\cite{mittal2012making} on real-world datasets, and LPIPS~\cite{zhang2018unreasonable} on paired datasets. 
% To evaluate the performance of different LLIE methods, we use both full-reference and non-reference image quality evaluation metrics. 
We also report PSNR and SSIM~\cite{wang2004image} as reference on paired datasets. 
% For real low-light image datasets without GT, we adopt two non-reference metrics, MUSIQ~\cite{ke2021musiq} and NIQE~\cite{mittal2012making}. 

\vspace{-1.5mm}
{\flushleft \bf Implementations.}
% \noindent{\bf Implementation Details.} 
Our VLM-IMI is based on SD2.1\footnote{https://huggingface.co/stabilityai/stable-diffusion-2-1}. We adopt LLaVA \cite{liu2023visual} to generate text instructions and select the text encoder of T5 \cite{raffel2020exploring} to obtain text embeddings as described in Sec \ref{sec:TGB}. The number of LIPF modules is $6$. The VLM-IMI is finetuned for 50k iterations with the AdamW \cite{kingma2014adam} optimizer on two NVIDIA RTX A6000 GPUs, where the batch size and learning rate are set to 8 and $5 \times 10^{-5}$, respectively. The training patch size is set as $512 \times 512$ resolution. For inference, we employ spaced DDPM sampling \cite{nichol2021improved} with 50 timesteps.

\begin{table*}[t]
% \vspace{-6.5mm}
\caption{Ablation study about fixed instruction while removing LIPF on LOL~\cite{wei2018deep}, LSRW~\cite{hai2023r2rnet}, DICM~\cite{lee2013contrast}, NPE~\cite{wang2013naturalness}, and VV~\cite{vonikakis2018evaluation}.}
\label{tab:removing IPFM_supp} 
\vspace{-2mm}
\centering
\scriptsize
\renewcommand{\arraystretch}{1}
\setlength{\tabcolsep}{0.8pt}
% 通过缩放适配页面宽度
%\resizebox{0.98\textwidth}{!}{
\begin{tabular}{l|ccc|c|ccc|ccc|cc|cc|cc}
\hline
\multirow{3}{*}{\textbf{ID}}&\multicolumn{3}{c|}{\textbf{Text Instructions}}& \textbf{Module}
& \multicolumn{3}{c}{\textbf{LOL} \cite{wei2018deep}} 
& \multicolumn{3}{c}{\textbf{LSRW} \cite{hai2023r2rnet}}
& \multicolumn{2}{c}{\textbf{DICM} \cite{lee2013contrast}}
& \multicolumn{2}{c}{\textbf{NPE} \cite{wang2013naturalness}}
& \multicolumn{2}{c}{\textbf{VV} \cite{vonikakis2018evaluation}}\\
\cline{2-17}
&\makecell {Lighting \\Information} & \makecell{Shadows \\ and Reflections} & \makecell{Spatial and \\ Contextual Information}
&LIPF&PSNR~$\uparrow$ & SSIM~$\uparrow$ & LPIPS~$\downarrow$
&PSNR~$\uparrow$ & SSIM~$\uparrow$ & LPIPS~$\downarrow$
&MUSIQ~$\uparrow$& NIQE~$\downarrow$
&MUSIQ~$\uparrow$& NIQE~$\downarrow$
&MUSIQ~$\uparrow$& NIQE~$\downarrow$
\\
\hline
\multirow{2}{*}{(a)}&\multirow{2}{*}{\ding{55}} & \multirow{2}{*}{\ding{55}} & \multirow{2}{*}{\ding{55}} & w/o &14.902	&0.587	&0.367&14.235	&0.284	&0.433	&57.702	&5.026	&57.937	&5.401&	38.827	&5.364 \\
&&&& w/& 18.137 & 0.694 & 0.263 
&17.924&	0.412&	0.281	&	59.853	&4.275	&60.124&	4.812	&42.187	&4.628\\
\hline
\multirow{2}{*}{(b)}&\multirow{2}{*}{\ding{51}} &\multirow{2}{*}{\ding{55}} & \multirow{2}{*}{\ding{55}} & w/o  &16.253	&0.650	&0.294	&	15.609	&0.377&	0.349	&	59.861&	4.463	&58.810&	4.928	&40.003	&5.027\\
&&&& w/& 20.336 & 0.726 & 0.191
&18.733	&0.499	&0.207	&61.372	&4.413	&63.001	&4.426	&43.692	&3.942\\
\hline
\multirow{2}{*}{(c)}&\multirow{2}{*}{\ding{51}} & \multirow{2}{*}{\ding{51}} & \multirow{2}{*}{\ding{55}} & w/o  &16.738	&0.629	&0.271	&16.227	&0.439	&0.278	&59.734	&4.608	&58.126	&4.763	&41.148	&4.825 \\
&&&& w/& 19.621 & 0.755 & 0.227
&19.063&	0.519	&0.214	&63.025	&3.916	&62.863	&3.875	&45.028	&4.071\\
\hline
\multirow{2}{*}{(d)}&\multirow{2}{*}{\ding{51}} & \multirow{2}{*}{\ding{51}} & \multirow{2}{*}{\ding{51}}& w/o  &18.534&0.715	&0.258	&17.419	&0.472	&0.267	&60.279	&4.057	&59.202	&4.637	&41.730	&4.508\\
&&&& w/ & \textbf{21.112} & \textbf{0.802} & \textbf{0.155} 
&\textbf{19.351}	&\textbf{0.548}	&\textbf{0.173}&\textbf{64.661}	&\textbf{3.551}&	\textbf{64.097}&	\textbf{3.421}&	\textbf{45.984}	&\textbf{3.523}

\\
\hline
\end{tabular}
%}
\end{table*}
%%%%%%%%%%%%%%%%%%%%%%
\begin{figure*}[t]
% \vspace{-3mm}
    \centering
    \includegraphics[width=\textwidth]{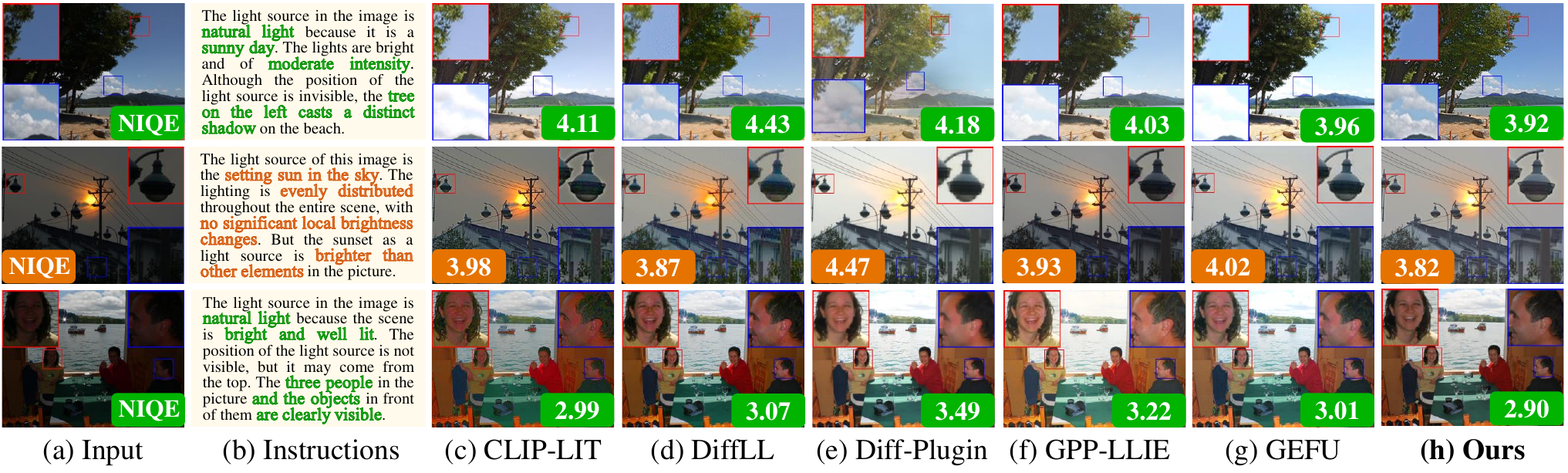}
   \vspace{-7mm}
    \caption{\textbf{Visual comparison on real datasets}, including DICM~\cite{lee2013contrast}, NPE~\cite{wang2013naturalness}, and VV~\cite{vonikakis2018evaluation}.
    Our VLM-IMI is able to generate results with better naturalness in terms of the realism metric, \ie, NIQE.
    }
    \label{fig:comparison_unpaired}
    % \vspace{-2mm}
\end{figure*}
\begin{figure*}[!t]
% \vspace{-9mm}
    \centering
    \begin{tabular}{c}
    \hspace{-2mm}\includegraphics[width=\textwidth]{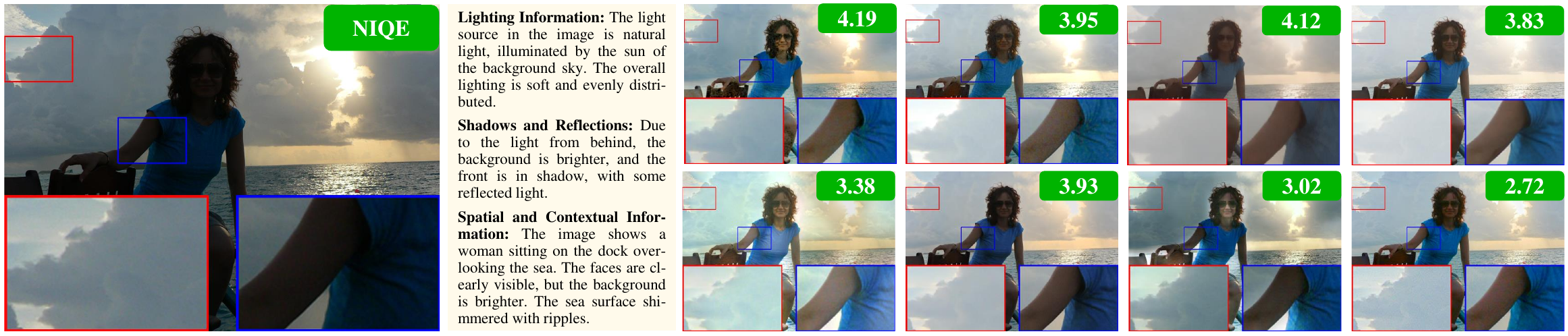}\\
    \small
    \makebox[0.27\textwidth][c]{(a) Input}
     \makebox[0.137\textwidth][c]{(b) Instructions (K=2)}
   \makebox[0.137\textwidth][c]{(c) Tab.~\ref{tab:removing IPFM_supp}(a)}
   \makebox[0.137\textwidth][c]{(d) Tab.~\ref{tab:removing IPFM_supp}(b)}
   \makebox[0.137\textwidth][c]{(e) Tab.~\ref{tab:removing IPFM_supp}(c)}
   \makebox[0.137\textwidth][c]{\textbf{(f) Ours}}
     \end{tabular}
     \vspace{-3mm}
    \caption{\textbf{Visual comparison} about text instructions on VV dataset~\cite{vonikakis2018evaluation}. Results without and with LIPF are presented in the top and bottom rows, respectively.
    Our full model generates results with better naturalness according to visual quality and the realism metric, NIQE.
    % We note that without using Lighting Information (b), Shadows and Reflections (c), and Spatial and Contextual Information (d), the model cannot produce visually-pleasing images.
    % In contrast, our full model is able to generate results with better naturalness according to visual quality as well as no-reference metrics, NIQE.
    }
    \label{fig:text prompts}
     \vspace{-3mm}
\end{figure*}
%%%%%%%%%%%%%%%%%%%%%
%\vspace{-1mm}
\subsection{Comparisons with state-of-the-arts}
%\vspace{-1mm}
% We compare our method with current representative low-light image enhancement approaches.
% , including traditional methods such as LIME \cite{guo2016lime}, CDEF \cite{lei2022low}, and BrainRetinex \cite{cai2023brain}, learning-based methods like RetinexNet \cite{wei2018deep}, DRBN \cite{yang2020fidelity}, Zero-DCE \cite{guo2020zero}, KinD++ \cite{zhang2021beyond}, EnlightenGAN \cite{jiang2021enlightengan}, LCDPNet \cite{wang2022local}, URetinexNet \cite{wu2022uretinex}, SCI \cite{ma2022toward}, BL \cite{ma2023bilevel}, PairLIE \cite{fu2023learning}, NeRCo \cite{yang2023implicit} and CLIP-LIT \cite{liang2023iterative}, as well as diffusion-based methods including GDP \cite{fei2023generative}, WeatherDiff \cite{ozdenizci2023restoring}, DiffLL \cite{jiang2023low}, GSAD \cite{hou2023global}, QuadPrior \cite{wang2024zero}. 
% It is important to note that our reported results represent the average of five independent evaluations.
%
% More results are shown in the Appendix.
\vspace{-1.5mm}
{\flushleft \bf Quantitative results.}
% \noindent{\bf Quantitative results.} 
Table \ref{tab:comparison} presents the quantitative results on paired LOL~\cite{wei2018deep} and LSRW~\cite{hai2023r2rnet} datasets. 
%
%Our method surpasses nearly all other methods, achieving SOTA performance. 
%
%Since DiffLL \cite{jiang2023low} and GSAD \cite{hou2023global} are trained in a fully supervised manner on the LOL dataset, their performance on it is better, but significantly degraded on unseen scenes. 
%
Compared with pre-trained models and methods trained on LOL, our VLM-IMI$^*$ achieves the best perception metric, \ie, LPIPS, on both LSRW and LOL datasets. 
%
% When trained on our dataset, VLM-IMI delivers the lowest LPIPS on the LOL. 
% On the LSRW, our VLM-IMI outperforms all others and achieves the highest values across all three metrics. 
%
In particular, the significant improvement in LPIPS provides compelling evidence for the superior perceptual quality of our method. 
Additionally, to further validate the effectiveness and robustness of our method, we conduct comparisons on three unpaired benchmarks: DICM~\cite{lee2013contrast}, NPE~\cite{wang2013naturalness}, and VV~\cite{vonikakis2018evaluation}. 
As shown in Table~\ref{tab:comparison}, our method achieves the highest MUSIQ score on all three datasets. 
For the NIQE score, we obtain the best result on the NPE and VV datasets and the second-best result on the DICM dataset. 
In summary, our method has better generalization ability than other competitive methods.

%\vspace{-1mm}
\vspace{-1.5mm}
{\flushleft \bf Visual comparisons.}
% \noindent{\bf Visual Comparisons.} 
Figure~\ref{fig:comparison_paired} presents visual comparisons on paired datasets: LOL~\cite{wei2018deep} and LSRW~\cite{hai2023r2rnet}. 
%Figure~\ref{fig:comparison_paired} presents a visual comparison on paired dataset LSRW~\cite{hai2023r2rnet}. 
%
We can see that previous methods yield results with incorrect exposure, color distortion, or noise amplification, while our method effectively enhances image contrast, suppresses noise, and reconstructs clearer details, producing perceptually superior results. 
Furthermore, Figure~\ref{fig:comparison_unpaired} illustrates the results on unpaired benchmarks: DICM~\cite{lee2013contrast}, NPE~\cite{wang2013naturalness}, and VV \cite{vonikakis2018evaluation}. 
Existing methods exhibit limited generalizability in these scenarios. 
Especially in row 1, most methods fail to effectively enhance illumination or suffer from overexposure, resulting in loss of image details. 
In contrast, our method effectively restores illumination, natural color, and clear details, demonstrating its strong generalizability to unseen scenes. 
% More comparisons are shown in the supplementary material.

%-------------------------------------------------------------------------
%\vspace{-1mm}
\subsection{Ablation studies}

\begin{table*}[!t]
% \vspace{-6.5mm}
\caption{Effect of LIPF on LOL~\cite{wei2018deep}, LSRW~\cite{hai2023r2rnet}, DICM~\cite{lee2013contrast}, NPE~\cite{wang2013naturalness}, and VV~\cite{vonikakis2018evaluation}.}
\label{tab:SPG_supp} 
\vspace{-2mm}
\centering
\scriptsize
\renewcommand{\arraystretch}{1}
\setlength{\tabcolsep}{3.6pt}
%\resizebox{0.5\textwidth}{!}{ % 缩放表格到90%的宽度
\begin{tabular}{l|l|c|c|ccc|ccc|cc|cc|cc}
\hline
\multirow{2}{*}{\textbf{ID}}
&\multirow{2}{*}{\textbf{Experiments}} 
&\multirow{2}{*}{\textbf{Arch}} 
&\multirow{2}{*}{\textbf{Norm}}
& \multicolumn{3}{c}{\textbf{LOL \cite{wei2018deep}}} 
& \multicolumn{3}{c}{\textbf{LSRW} \cite{hai2023r2rnet}}
& \multicolumn{2}{c}{\textbf{DICM} \cite{lee2013contrast}}
& \multicolumn{2}{c}{\textbf{NPE} \cite{wang2013naturalness}}
& \multicolumn{2}{c}{\textbf{VV} \cite{vonikakis2018evaluation}}
\\
\cline{5-16}
&&&
&PSNR~$\uparrow$ & SSIM~$\uparrow$ & LPIPS~$\downarrow$
&PSNR~$\uparrow$ & SSIM~$\uparrow$ & LPIPS~$\downarrow$
&MUSIQ~$\uparrow$& NIQE~$\downarrow$
&MUSIQ~$\uparrow$& NIQE~$\downarrow$
&MUSIQ~$\uparrow$& NIQE~$\downarrow$ \\
\hline
(a)& w/o LIPF & None & None & 18.534 & 0.715 & 0.258 
&17.419	&0.472	&0.267	&60.279	&4.057	&59.202	&4.637	&41.730	&4.508\\
(b)& w/o LIPF & MLP & LN & 19.376 & 0.761 & 0.261
&18.326&	0.501	&0.228	&62.351	&4.192	&61.733	&4.302	&43.095	&4.061\\
(c)& w/o timesteps & LIPF & LN & 20.680 & 0.787 & 0.197 
&19.011	&0.532&	0.202	&63.904	&3.836&	62.881	&3.719	&44.931	&3.867\\
\hline
\textbf{(d)}& \textbf{Ours }& LIPF & AdaLN & \textbf{21.112} & \textbf{0.802} & \textbf{0.155} 
&\textbf{19.351}	&\textbf{0.548}	&\textbf{0.173}	&\textbf{64.661}	&\textbf{3.551}	&\textbf{64.097}	&\textbf{3.421}	&\textbf{45.984}&\textbf{3.523}\\
\hline
\end{tabular}
%}
\vspace{-2mm}
\end{table*}
\begin{figure*}[!t]
% \vspace{-9mm}
    \centering
    \begin{tabular}{c}
    \hspace{-3mm}\includegraphics[width=\textwidth]{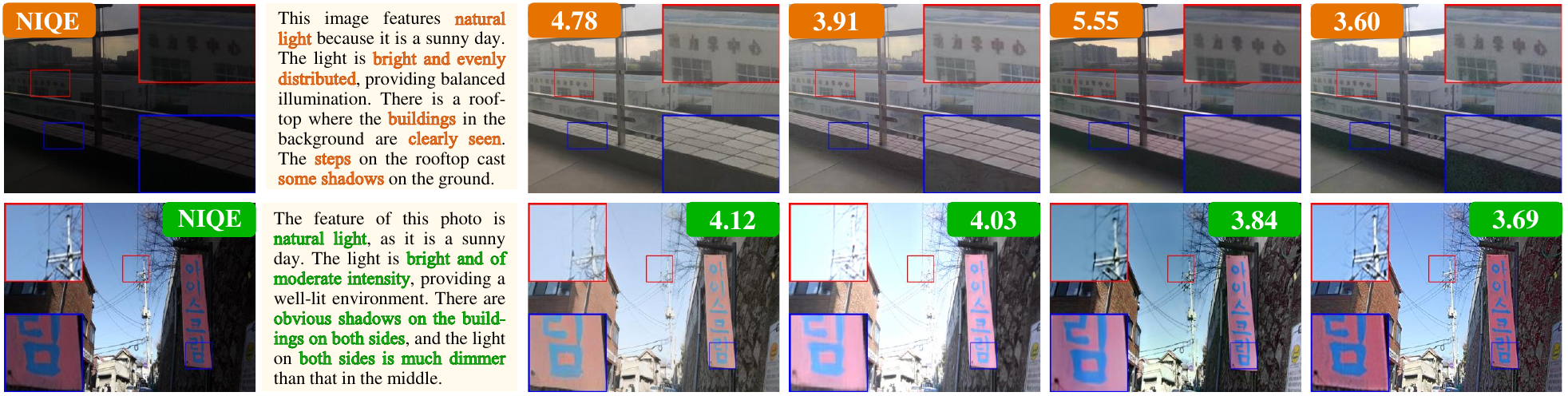}\\
    \small
    \makebox[0.16\textwidth][c]{(a) Input}
     \makebox[0.16\textwidth][c]{(b) Instructions}
   \makebox[0.163\textwidth][c]{(c) Tab.~\ref{tab:SPG_supp}(a)}
   \makebox[0.163\textwidth][c]{(d) Tab.~\ref{tab:SPG_supp}(b)}
   \makebox[0.163\textwidth][c]{(e) Tab.~\ref{tab:SPG_supp}(c)}
   \makebox[0.163\textwidth][c]{\textbf{(f) Ours}}
     \end{tabular}
     \vspace{-3mm}
    \caption{\textbf{Visual comparison} about LIPF on LSRW~\cite{hai2023r2rnet} and DICM~\cite{lee2013contrast} datasets. 
    Our VLM-IMI generates more natural and vivid results.
    }
    \label{fig:ab_LIPF}
    \vspace{-3mm}
\end{figure*}
%%%%%%%%%%%%%%%%%%%%
\begin{figure}[!t]
% \vspace{2mm}
    \centering
    \begin{tabular}{ccc}
    \hspace{-2mm}\includegraphics[width=0.323\linewidth]{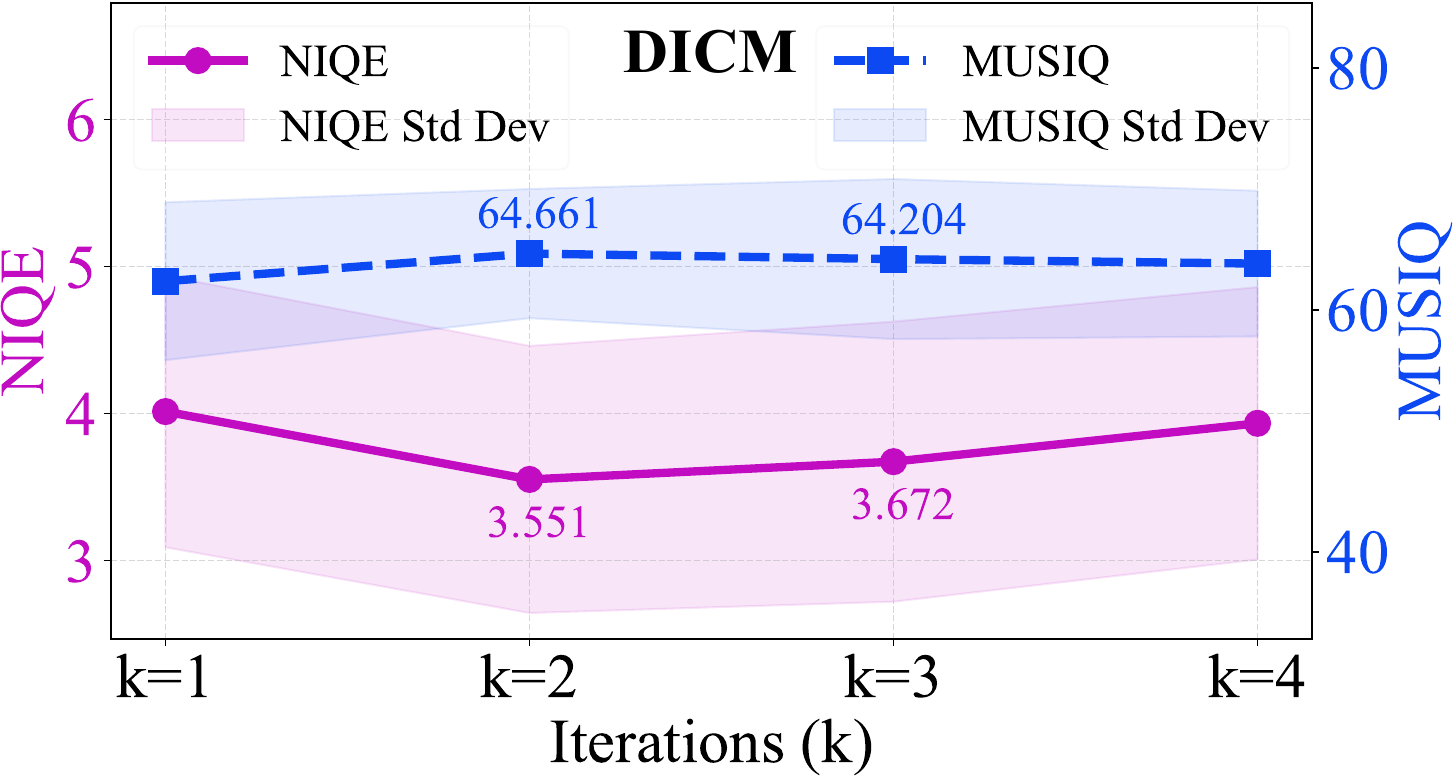}&
    \hspace{-3mm}\includegraphics[width=0.323\linewidth]{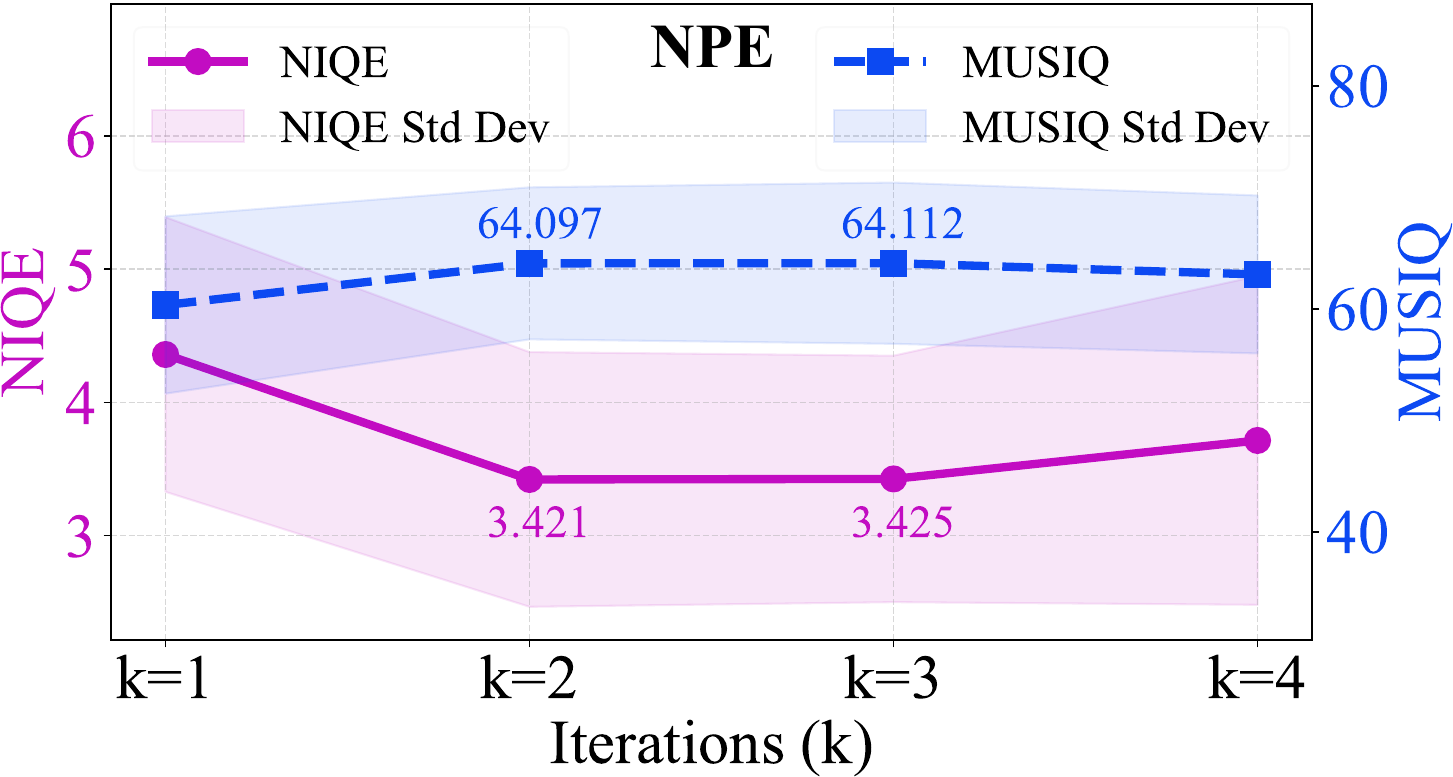}&
    \hspace{-3mm}\includegraphics[width=0.323\linewidth]{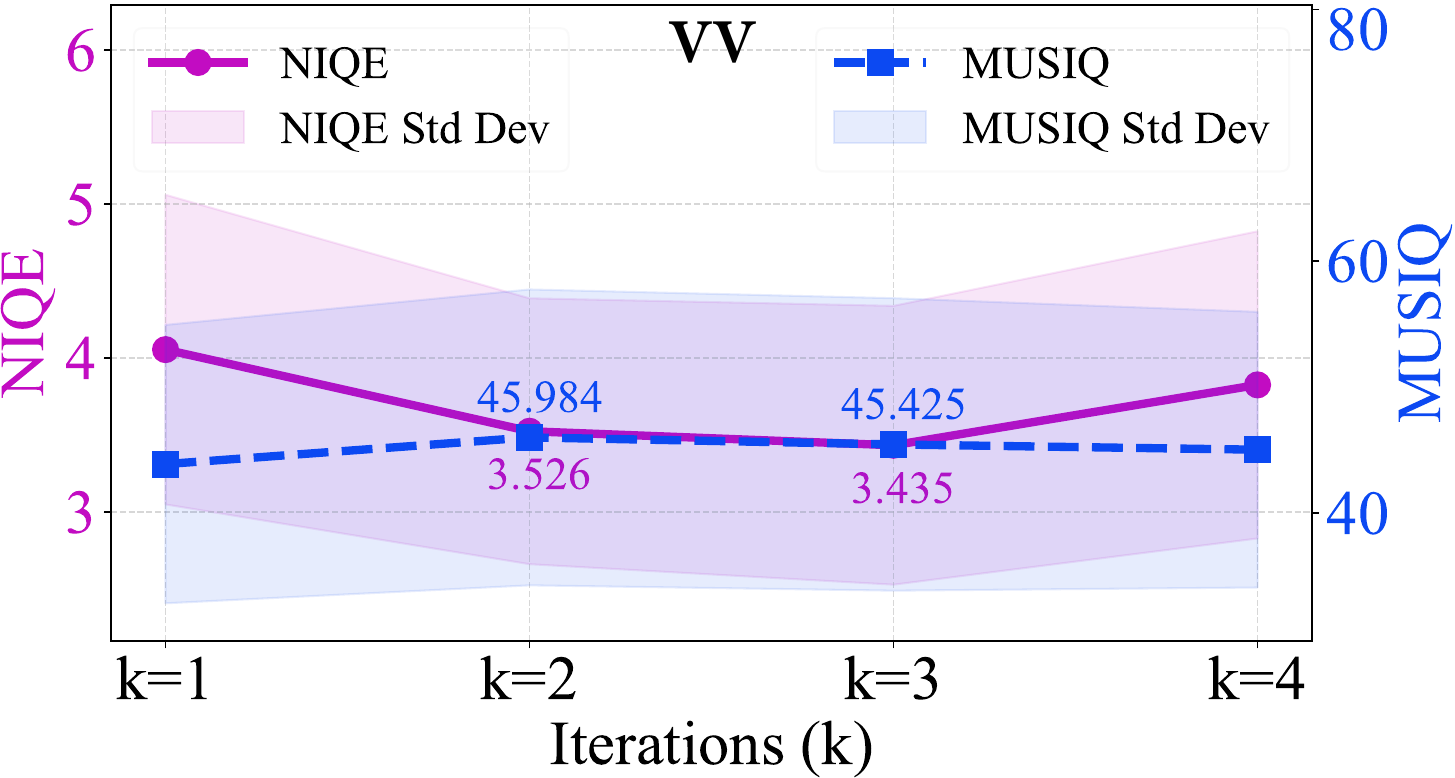}
     \end{tabular}
     \vspace{-2mm}
    \caption{\textbf{Effect of iterative instruction strategy} on three real-world datasets. 
    The results are the best at the second iteration.
    % Blue lines represent NIQE scores (left y-axis), and orange lines denote MUSIQ scores (right y-axis). 
    % Shaded regions indicate the standard deviation for each metric.
    }
     \label{fig:NIQE vs MUSIQ}
\vspace{-2mm}
\end{figure}
%%%%%%%%%%%%%%%%%%%%%%
% \begin{figure}[!t]
% % \vspace{2mm}
%     \centering
%     \begin{tabular}{c}
%     \hspace{-2mm}\includegraphics[width=\linewidth]{LaTeX/images/Figure9.pdf}
%     \end{tabular}
%     \vspace{-2mm}
%        \caption{\textbf{Visual comparison of iterative instruction strategy} on DICM \cite{lee2013contrast}.
%      We note that the results improve with the second iteration, while the first iteration leads to under-exposure results, and more iterations produce over-exposure results.
%     }
%      \label{fig:NIQE vs MUSIQ}
% \vspace{-5mm}
% \end{figure}

\vspace{-1.5mm}
{\flushleft \bf Effect on design of text instructions.}
% \noindent{\bf Effect on Design of Text Instructions.} 
To further explore the contribution of various types of instructions to enhanced performance, we design and conduct ablation experiments on text instructions. The experimental results are shown in Table \ref{tab:removing IPFM_supp} and Figure \ref{fig:text prompts}. It can be observed that the absence of diverse types of text instructions leads to a noticeable degradation in both the quantitative and visual results, thereby impairing the model's effectiveness in restoring the details and illumination of low-light images.  %More visual results are given in the supplementary material.
% Moreover, using LIPF consistently improves the enhancement quality on various text instructions compared with the model without LIPF, indicating the effectiveness of LIPF.
Moreover, incorporating LIPF consistently enhances the image quality across diverse textual instructions, demonstrating the effectiveness of the proposed LIPF.

\vspace{-1.5mm}
{\flushleft \bf Effect on learnable instruction prior fusion.}
% \noindent{\bf Effect of Instruction Prior Fusion Module.} 
To validate the effectiveness of the learnable instruction prior fusion module, we conduct experiments under four settings: (a) removing LIPF, (b) replacing the LIPF blocks with MLP layers, (c) replacing AdaLN with LN to remove timesteps, and (d) our setting. As shown in Table \ref{tab:SPG_supp} and Figure~\ref{fig:ab_LIPF}, removing LIPF significantly degrades performance, particularly in complex scenes where the model struggles to capture lighting and spatial structures. 
% \begin{wraptable}{r}{0.5\textwidth}
% \begin{table}
% \caption{Ablation study on IPFM.}
% \label{tab:SPG} % 设置引用表格的标识符
% \centering
% \resizebox{0.5\textwidth}{!}{ % 缩放表格到90%的宽度
% \begin{tabular}{l|l|c|c|ccc}
% \hline
% \multirow{2}{*}{ID}&\multirow{2}{*}{Experiments} & \multirow{2}{*}{Arch} & \multirow{2}{*}{Norm}& \multicolumn{3}{c}{LOL \cite{wei2018deep}}
% \\
% \cline{5-7}
% &&&& PSNR~$\uparrow$ & SSIM~$\uparrow$ & LPIPS~$\downarrow$ \\
% \hline
% (a)& w/o IPFM & None & None & 18.534 & 0.715 & 0.258 \\
% (b)& w/o IPFM & MLP & LN & 19.376 & 0.761 & 0.261 \\
% (c)& w/o timesteps & IPFM & LN & 20.680 & 0.787 & 0.197 \\
% \hline
% (d)& Ours & IPFM & AdaLN & \textbf{21.112} & \textbf{0.802} & \textbf{0.155} \\
% \hline
% \end{tabular}
% }
% \end{table}
% \end{wraptable}
Moreover, our LIPF modules outperform MLP layers in leveraging LLM capabilities for diffusion models, enhancing global dependencies and semantic understanding. Integrating timestep information via AdaLN further improves noise prediction by dynamically adapting text features at different diffusion stages, which is crucial for optimizing the recovery process.

\begin{figure}[t]
% \vspace{2mm}
    \centering
    \begin{tabular}{c}
\hspace{-2mm}\includegraphics[width=0.99\linewidth]{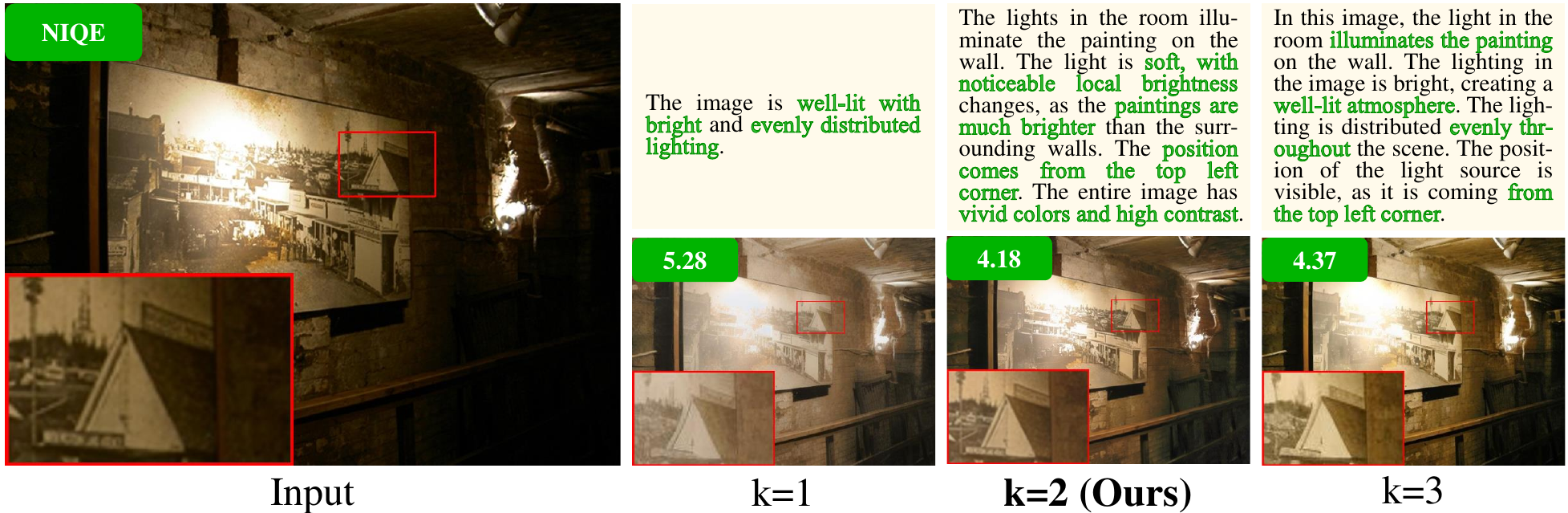}\\
     \end{tabular}
     \vspace{-3mm}
    \caption{\textbf{Visual comparison of iterative instruction strategy} on DICM \cite{lee2013contrast}.
    We note that the results improve with the second iteration, while the first iteration leads to under-exposure results, and more iterations produce over-exposure results.
    }
     \label{fig:feedback}
\vspace{-3mm}
\end{figure}

\vspace{-1.5mm}
{\flushleft \bf Effect on inference with iterative instruction strategy.}
% \noindent{\bf Effect of Iterative Instruction Strategy.} 
The effectiveness of the inference with the iterative instruction strategy is demonstrated in Figure~\ref{fig:NIQE vs MUSIQ} and Figure~\ref{fig:feedback}. By comparing image quality across different iterations, we find that $\mathit{k}=2$ is sufficient to achieve high-quality enhancement results. After customized instruction, the contrast and brightness in the shadow areas significantly improve, with finer details becoming more pronounced and the overall appearance appearing more natural and realistic.

\vspace{-1.5mm}
{\flushleft \bf Effect on inference with manual instruction strategy.}
All of Figure~\ref{fig:pipeline}(e) and Figure~\ref{fig:controllability} exhibit that our VLM-IMI's output can be controlled by manual instructions.
This approach also suggests that our VLM-IMI can adjust the results according to different lighting scenarios to flexibly handle different images, thereby effectively improving visual performance.

\vspace{-1.5mm}
%%%%%%%%%%%%%%%%%
\section{Conclusion}
% \vspace{-1.5mm}
% We have proposed VLM-IMI, a multimodal instruction learning method for low-light image enhancement. 
% % %
% We explore the role of different instruction priors in LLIE and design instruction templates to guide VLM in generating text instructions as accurate cross-modal instruction priors. 
% %
% To facilitate the integration of instruction priors, we introduce an instruction prior fusion module that dynamically interacts with text features, improving guidance for noise prediction at each diffusion stage. 
% %
% Additionally, an iterative and manual instruction strategy is proposed to handle real low-light images with different light scenarios. 
% %
% Extensive experimental results demonstrate that our VLM-IMI outperforms state-of-the-art LLIE methods.
We have proposed VLM-IMI, a multimodal instruction learning framework designed for LLIE. Our approach investigates the role of diverse instruction priors in guiding enhancement and introduces carefully designed instruction templates to enable VLMs to generate accurate cross-modal guidance. 
To effectively integrate these priors, we develop a learnable instruction prior fusion module, which dynamically interacts with textual features across diffusion stages, enhancing the model’s ability to predict noise and preserve semantic fidelity. 
Furthermore, we introduce an iterative and manual instruction strategy to adaptively refine enhancement results for real-world low-light images during inference. 
Extensive experiments demonstrate that VLM-IMI outperforms state-of-the-art LLIE methods in terms of perception and realism on diverse lighting conditions.
%
% With a unique vision-language integration mechanism, our framework presents a new paradigm for text-driven image enhancement, expanding the capabilities of AI-assisted low-light restoration techniques.

\vspace{-1.5mm}
{\flushleft \bf Acknowledgments:} This work is supported in part by the National Natural Science Foundation of China No.62476041.

{
    \small
    \bibliographystyle{ieeenat_fullname}
    \bibliography{main}

@String(IJCV = {Int. J. Comput. Vis.})

@String(CVPR= {IEEE Conf. Comput. Vis. Pattern Recog.})

@String(ICCV= {Int. Conf. Comput. Vis.})

@String(ECCV= {Eur. Conf. Comput. Vis.})

@String(TOG= {ACM Trans. Graph.})

@String(TIP  = {IEEE Trans. Image Process.})

@String(TMM  = {IEEE Trans. Multimedia})

@String(IJCAI = {IJCAI})

@String(PR   = {Pattern Recognition})

@String(AAAI = {AAAI})

@String(SPL	= {IEEE Sign. Process. Letters})

@String(IJCV  = {IJCV})

@String(CVPR  = {CVPR})

@String(ICCV  = {ICCV})

@String(ECCV  = {ECCV})

@String(TOG   = {ACM TOG})

@String(TIP   = {IEEE TIP})

@String(TCSVT = {IEEE TCSVT})

@String(TMM   =	{IEEE TMM})

@String(PR = {PR})

@article{land1977retinex,
  title={The retinex theory of color vision},
  author={Land, Edwin H},
  journal={SciAm},
  year={1977},
}

@article{wang2004image,
  title={Image quality assessment: from error visibility to structural similarity},
  author={Wang, Zhou and Bovik, Alan C and Sheikh, Hamid R and Simoncelli, Eero P},
  journal={{IEEE} TIP},
  year={2004},
}

@book{bovik2010handbook,
  title={Handbook of image and video processing},
  author={Bovik, Alan C},
  year={2010},
  publisher={Academic press}
}

@article{mittal2012making,
  title={Making a “completely blind” image quality analyzer},
  author={Mittal, Anish and Soundararajan, Rajiv and Bovik, Alan C},
  journal={{IEEE} SPL},
  year={2012},
}

@article{lee2013contrast,
  title={Contrast enhancement based on layered difference representation of 2D histograms},
  author={Lee, Chulwoo and Lee, Chul and Kim, Chang-Su},
  journal={{IEEE} TIP},
  year={2013},
}

@article{wang2013naturalness,
  title={Naturalness preserved enhancement algorithm for non-uniform illumination images},
  author={Wang, Shuhang and Zheng, Jin and Hu, Hai-Miao and Li, Bo},
  journal={{IEEE} TIP},
  year={2013}
}

@article{kingma2014adam,
  title={Adam: A method for stochastic optimization},
  author={Kingma, Diederik P and Ba, Jimmy},
  journal={arXiv preprint arXiv:1412.6980},
  year={2014}
}

@inproceedings{dang2015raise,
  title={Raise: A raw images dataset for digital image forensics},
  author={Dang-Nguyen, Duc-Tien and Pasquini, Cecilia and Conotter, Valentina and Boato, Giulia},
  booktitle={ACM MMSys},
  year={2015}
}

@article{guo2016lime,
  title={LIME: Low-light image enhancement via illumination map estimation},
  author={Guo, Xiaojie and Li, Yu and Ling, Haibin},
  journal={{IEEE} TIP},
  year={2016},
}

@inproceedings{lin2017feature,
  title={Feature pyramid networks for object detection},
  author={Lin, Tsung-Yi and Doll{\'a}r, Piotr and Girshick, Ross and He, Kaiming and Hariharan, Bharath and Belongie, Serge},
  booktitle={CVPR},
  year={2017}
}

@article{lore2017llnet,
  title={LLNet: A deep autoencoder approach to natural low-light image enhancement},
  author={Lore, Kin Gwn and Akintayo, Adedotun and Sarkar, Soumik},
  journal={PR},
  year={2017},
}

@inproceedings{zhang2018unreasonable,
  title={The unreasonable effectiveness of deep features as a perceptual metric},
  author={Zhang, Richard and Isola, Phillip and Efros, Alexei A and Shechtman, Eli and Wang, Oliver},
  booktitle={CVPR},
  year={2018}
}

@article{vonikakis2018evaluation,
  title={On the evaluation of illumination compensation algorithms},
  author={Vonikakis, Vassilios and Kouskouridas, Rigas and Gasteratos, Antonios},
  journal={MTAP},
  year={2018},
}

@article{li2018structure,
  title={Structure-revealing low-light image enhancement via robust retinex model},
  author={Li, Mading and Liu, Jiaying and Yang, Wenhan and Sun, Xiaoyan and Guo, Zongming},
  journal={{IEEE} TIP},
  year={2018},
}

@article{wei2018deep,
  title={Deep retinex decomposition for low-light enhancement},
  author={Wei, Chen and Wang, Wenjing and Yang, Wenhan and Liu, Jiaying},
  journal={arXiv preprint arXiv:1808.04560},
  year={2018}
}

@article{lee2018pre,
  title={Pre-training of deep bidirectional transformers for language understanding},
  author={Lee, JDMCK and Toutanova, K},
  journal={arXiv preprint arXiv:1810.04805},
  year={2018}
}

@inproceedings{deng2019arcface,
  title={Arcface: Additive angular margin loss for deep face recognition},
  author={Deng, Jiankang and Guo, Jia and Xue, Niannan and Zafeiriou, Stefanos},
  booktitle={CVPR},
  year={2019}
}

@inproceedings{wang2019underexposed,
  title={Underexposed photo enhancement using deep illumination estimation},
  author={Wang, Ruixing and Zhang, Qing and Fu, Chi-Wing and Shen, Xiaoyong and Zheng, Wei-Shi and Jia, Jiaya},
  booktitle={CVPR},
  year={2019}
}

@inproceedings{guo2020zero,
  title={Zero-reference deep curve estimation for low-light image enhancement},
  author={Guo, Chunle and Li, Chongyi and Guo, Jichang and Loy, Chen Change and Hou, Junhui and Kwong, Sam and Cong, Runmin},
  booktitle={CVPR},
  year={2020}
}

@article{raffel2020exploring,
  title={Exploring the limits of transfer learning with a unified text-to-text transformer},
  author={Raffel, Colin and Shazeer, Noam and Roberts, Adam and Lee, Katherine and Narang, Sharan and Matena, Michael and Zhou, Yanqi and Li, Wei and Liu, Peter J},
  journal={JMLR},
  year={2020}
}

@article{ho2020denoising,
  title={Denoising diffusion probabilistic models},
  author={Ho, Jonathan and Jain, Ajay and Abbeel, Pieter},
  journal={NeurIPS},
  year={2020}
}

@inproceedings{yang2020fidelity,
  title={From fidelity to perceptual quality: A semi-supervised approach for low-light image enhancement},
  author={Yang, Wenhan and Wang, Shiqi and Fang, Yuming and Wang, Yue and Liu, Jiaying},
  booktitle={CVPR},
  year={2020}
}

@article{xu2021exploring,
  title={Exploring image enhancement for salient object detection in low light images},
  author={Xu, Xin and Wang, Shiqin and Wang, Zheng and Zhang, Xiaolong and Hu, Ruimin},
  journal={{ACM} TOMM},
  year={2021},
}

@article{zhang2021beyond,
  title={Beyond brightening low-light images},
  author={Zhang, Yonghua and Guo, Xiaojie and Ma, Jiayi and Liu, Wei and Zhang, Jiawan},
  journal={IJCV},
  year={2021},
}

@article{jiang2021enlightengan,
  title={Enlightengan: Deep light enhancement without paired supervision},
  author={Jiang, Yifan and Gong, Xinyu and Liu, Ding and Cheng, Yu and Fang, Chen and Shen, Xiaohui and Yang, Jianchao and Zhou, Pan and Wang, Zhangyang},
  journal={{IEEE} TIP},
  year={2021},
}

@inproceedings{ke2021musiq,
  title={Musiq: Multi-scale image quality transformer},
  author={Ke, Junjie and Wang, Qifei and Wang, Yilin and Milanfar, Peyman and Yang, Feng},
  booktitle={ICCV},
  year={2021}
}

@inproceedings{radford2021learning,
  title={Learning transferable visual models from natural language supervision},
  author={Radford, Alec and Kim, Jong Wook and Hallacy, Chris and Ramesh, Aditya and Goh, Gabriel and Agarwal, Sandhini and Sastry, Girish and Askell, Amanda and Mishkin, Pamela and Clark, Jack and others},
  booktitle={ICML},
  year={2021},
}

@inproceedings{nichol2021improved,
  title={Improved denoising diffusion probabilistic models},
  author={Nichol, Alexander Quinn and Dhariwal, Prafulla},
  booktitle={ICML},
  year={2021},
}

@article{alayrac2022flamingo,
  title={Flamingo: a visual language model for few-shot learning},
  author={Alayrac, Jean-Baptiste and Donahue, Jeff and Luc, Pauline and Miech, Antoine and Barr, Iain and Hasson, Yana and Lenc, Karel and Mensch, Arthur and Millican, Katherine and Reynolds, Malcolm and others},
  journal={NeurIPS},
  year={2022}
}

@inproceedings{li2022blip,
  title={Blip: Bootstrapping language-image pre-training for unified vision-language understanding and generation},
  author={Li, Junnan and Li, Dongxu and Xiong, Caiming and Hoi, Steven},
  booktitle={ICML},
  year={2022},
}

@article{yang2022rethinking,
  title={Rethinking low-light enhancement via transformer-GAN},
  author={Yang, Shaoliang and Zhou, Dongming and Cao, Jinde and Guo, Yanbu},
  journal={{IEEE} SPL},
  year={2022},
}

@inproceedings{wang2022local,
  title={Local color distributions prior for image enhancement},
  author={Wang, Haoyuan and Xu, Ke and Lau, Rynson WH},
  booktitle={ECCV},
  year={2022},
}

@article{lei2022low,
  title={Low-light image enhancement using the cell vibration model},
  author={Lei, Xiaozhou and Fei, Zixiang and Zhou, Wenju and Zhou, Huiyu and Fei, Minrui},
  journal={{IEEE} TMM},
  year={2022},
}

@inproceedings{lugmayr2022repaint,
  title={Repaint: Inpainting using denoising diffusion probabilistic models},
  author={Lugmayr, Andreas and Danelljan, Martin and Romero, Andres and Yu, Fisher and Timofte, Radu and Van Gool, Luc},
  booktitle={CVPR},
  year={2022}
}

@inproceedings{rombach2022high,
  title={High-resolution image synthesis with latent diffusion models},
  author={Rombach, Robin and Blattmann, Andreas and Lorenz, Dominik and Esser, Patrick and Ommer, Bj{\"o}rn},
  booktitle={CVPR},
  year={2022}
}

@inproceedings{wu2022uretinex,
  title={Uretinex-net: Retinex-based deep unfolding network for low-light image enhancement},
  author={Wu, Wenhui and Weng, Jian and Zhang, Pingping and Wang, Xu and Yang, Wenhan and Jiang, Jianmin},
  booktitle={CVPR},
  year={2022}
}

@inproceedings{ma2022toward,
  title={Toward fast, flexible, and robust low-light image enhancement},
  author={Ma, Long and Ma, Tengyu and Liu, Risheng and Fan, Xin and Luo, Zhongxuan},
  booktitle={CVPR},
  year={2022}
}

@inproceedings{whang2022deblurring,
  title={Deblurring via stochastic refinement},
  author={Whang, Jay and Delbracio, Mauricio and Talebi, Hossein and Saharia, Chitwan and Dimakis, Alexandros G and Milanfar, Peyman},
  booktitle={CVPR},
  year={2022}
}

@inproceedings{cai2023retinexformer,
  title={Retinexformer: One-stage retinex-based transformer for low-light image enhancement},
  author={Cai, Yuanhao and Bian, Hao and Lin, Jing and Wang, Haoqian and Timofte, Radu and Zhang, Yulun},
  booktitle={ICCV},
  year={2023}
}

@inproceedings{fu2023learning,
  title={Learning a simple low-light image enhancer from paired low-light instances},
  author={Fu, Zhenqi and Yang, Yan and Tu, Xiaotong and Huang, Yue and Ding, Xinghao and Ma, Kai-Kuang},
  booktitle={CVPR},
  year={2023}
}

@article{ma2023bilevel,
  title={Bilevel fast scene adaptation for low-light image enhancement},
  author={Ma, Long and Jin, Dian and An, Nan and Liu, Jinyuan and Fan, Xin and Luo, Zhongxuan and Liu, Risheng},
  journal={IJCV},
  year={2023},
}

@article{cheng2023black,
  title={Black-box prompt optimization: Aligning large language models without model training},
  author={Cheng, Jiale and Liu, Xiao and Zheng, Kehan and Ke, Pei and Wang, Hongning and Dong, Yuxiao and Tang, Jie and Huang, Minlie},
  journal={arXiv preprint arXiv:2311.04155},
  year={2023}
}

@article{luo2023controlling,
  title={Controlling vision-language models for universal image restoration},
  author={Luo, Ziwei and Gustafsson, Fredrik K and Zhao, Zheng and Sj{\"o}lund, Jens and Sch{\"o}n, Thomas B},
  journal={arXiv preprint arXiv:2310.01018},
  year={2023}
}

@article{hou2023global,
  title={Global structure-aware diffusion process for low-light image enhancement},
  author={Hou, Jinhui and Zhu, Zhiyu and Hou, Junhui and Liu, Hui and Zeng, Huanqiang and Yuan, Hui},
  journal={NeurIPS},
  year={2023}
}

@article{jiang2023low,
  title={Low-light image enhancement with wavelet-based diffusion models},
  author={Jiang, Hai and Luo, Ao and Fan, Haoqiang and Han, Songchen and Liu, Shuaicheng},
  journal={{ACM} TOG},
  year={2023},
}

@article{cai2023brain,
  title={Brain-like retinex: A biologically plausible retinex algorithm for low light image enhancement},
  author={Cai, Rongtai and Chen, Zekun},
  journal={PR},
  year={2023},
}

@inproceedings{yi2023diff,
  title={Diff-retinex: Rethinking low-light image enhancement with a generative diffusion model},
  author={Yi, Xunpeng and Xu, Han and Zhang, Hao and Tang, Linfeng and Ma, Jiayi},
  booktitle={ICCV},
  year={2023}
}

@article{hai2023r2rnet,
  title={R2rnet: Low-light image enhancement via real-low to real-normal network},
  author={Hai, Jiang and Xuan, Zhu and Yang, Ren and Hao, Yutong and Zou, Fengzhu and Lin, Fang and Han, Songchen},
  journal={JVCIR},
  year={2023},
}

@inproceedings{fei2023generative,
  title={Generative diffusion prior for unified image restoration and enhancement},
  author={Fei, Ben and Lyu, Zhaoyang and Pan, Liang and Zhang, Junzhe and Yang, Weidong and Luo, Tianyue and Zhang, Bo and Dai, Bo},
  booktitle={CVPR},
  year={2023}
}

@inproceedings{yang2023implicit,
  title={Implicit neural representation for cooperative low-light image enhancement},
  author={Yang, Shuzhou and Ding, Moxuan and Wu, Yanmin and Li, Zihan and Zhang, Jian},
  booktitle={ICCV},
  year={2023}
}

@inproceedings{liang2023iterative,
  title={Iterative prompt learning for unsupervised backlit image enhancement},
  author={Liang, Zhexin and Li, Chongyi and Zhou, Shangchen and Feng, Ruicheng and Loy, Chen Change},
  booktitle={ICCV},
  year={2023}
}

@article{ozdenizci2023restoring,
  title={Restoring vision in adverse weather conditions with patch-based denoising diffusion models},
  author={{\"O}zdenizci, Ozan and Legenstein, Robert},
  journal={IEEE TPAMI},
  year={2023},
}

@inproceedings{ruiz2023dreambooth,
  title={Dreambooth: Fine tuning text-to-image diffusion models for subject-driven generation},
  author={Ruiz, Nataniel and Li, Yuanzhen and Jampani, Varun and Pritch, Yael and Rubinstein, Michael and Aberman, Kfir},
  booktitle={CVPR},
  year={2023}
}

@inproceedings{zhang2023adding,
  title={Adding conditional control to text-to-image diffusion models},
  author={Zhang, Lvmin and Rao, Anyi and Agrawala, Maneesh},
  booktitle={ICCV},
  year={2023}
}

@article{touvron2023llama,
  title={Llama: Open and efficient foundation language models},
  author={Touvron, Hugo and Lavril, Thibaut and Izacard, Gautier and Martinet, Xavier and Lachaux, Marie-Anne and Lacroix, Timoth{\'e}e and Rozi{\`e}re, Baptiste and Goyal, Naman and Hambro, Eric and Azhar, Faisal and others},
  journal={arXiv preprint arXiv:2302.13971},
  year={2023}
}

@article{liu2023visual,
  title={Visual instruction tuning},
  author={Liu, Haotian and Li, Chunyuan and Wu, Qingyang and Lee, Yong Jae},
  journal={NeurIPS},
  year={2023}
}

@article{liu2023accelerating,
  title={Accelerating diffusion models for inverse problems through shortcut sampling},
  author={Liu, Gongye and Sun, Haoze and Li, Jiayi and Yin, Fei and Yang, Yujiu},
  journal={arXiv preprint arXiv:2305.16965},
  year={2023}
}

@inproceedings{wang2024zero,
  title={Zero-reference low-light enhancement via physical quadruple priors},
  author={Wang, Wenjing and Yang, Huan and Fu, Jianlong and Liu, Jiaying},
  booktitle={CVPR},
  year={2024}
}

@inproceedings{wu2024seesr,
  title={Seesr: Towards semantics-aware real-world image super-resolution},
  author={Wu, Rongyuan and Yang, Tao and Sun, Lingchen and Zhang, Zhengqiang and Li, Shuai and Zhang, Lei},
  booktitle={CVPR},
  year={2024}
}

@inproceedings{qu2024xpsr,
  title={Xpsr: Cross-modal priors for diffusion-based image super-resolution},
  author={Qu, Yunpeng and Yuan, Kun and Zhao, Kai and Xie, Qizhi and Hao, Jinhua and Sun, Ming and Zhou, Chao},
  booktitle={ECCV},
  year={2024},
}

@inproceedings{liu2024diff,
  title={Diff-plugin: Revitalizing details for diffusion-based low-level tasks},
  author={Liu, Yuhao and Ke, Zhanghan and Liu, Fang and Zhao, Nanxuan and Lau, Rynson WH},
  booktitle={CVPR},
  year={2024}
}

@inproceedings{li2024light,
  title={Light the night: A multi-condition diffusion framework for unpaired low-light enhancement in autonomous driving},
  author={Li, Jinlong and Li, Baolu and Tu, Zhengzhong and Liu, Xinyu and Guo, Qing and Juefei-Xu, Felix and Xu, Runsheng and Yu, Hongkai},
  booktitle={CVPR},
  year={2024}
}

@article{guo2025compressionawareonestepdiffusionmodel,
      title={Compression-Aware One-Step Diffusion Model for JPEG Artifact Removal}, 
      author={Jinpei Guo and Zheng Chen and Wenbo Li and Yong Guo and Yulun Zhang},
      year={2025},
      journal={arXiv preprint arXiv:2502.09873},    
}

@inproceedings{2017Grad,
  title={Grad-CAM: Visual Explanations from Deep Networks via Gradient-Based Localization},
  author={ Selvaraju, Ramprasaath R.  and  Cogswell, Michael  and  Das, Abhishek  and  Vedantam, Ramakrishna  and  Parikh, Devi  and  Batra, Dhruv },
  booktitle={ICCV},
  year={2017},
}

@inproceedings{InstructIR24,
  author       = {Marcos V. Conde and
                  Gregor Geigle and
                  Radu Timofte},
  title        = {InstructIR: High-Quality Image Restoration Following Human Instructions},
  booktitle    = {ECCV},
  year         = {2024},
}

@inproceedings{diffplugin24,
  author       = {Yuhao Liu and
                  Zhanghan Ke and
                  Fang Liu and
                  Nanxuan Zhao and
                  Rynson W. H. Lau},
  title        = {Diff-Plugin: Revitalizing Details for Diffusion-Based Low-Level Tasks},
  booktitle    = {CVPR},
  year         = {2024},
}

@inproceedings{UniProcessor24,
  author       = {Huiyu Duan and
                  Xiongkuo Min and
                  Sijing Wu and
                  Wei Shen and
                  Guangtao Zhai},
  title        = {UniProcessor: {A} Text-Induced Unified Low-Level Image Processor},
  booktitle    = {ECCV},
  year         = {2024},
}

@inproceedings{GPP-LLIE25,
  author       = {Han Zhou and
                  Wei Dong and
                  Xiaohong Liu and
                  Yulun Zhang and
                  Guangtao Zhai and
                  Jun Chen},
  title        = {Low-Light Image Enhancement via Generative Perceptual Priors},
  booktitle    = {AAAI},
  year         = {2025},
}

@inproceedings{GEFU25,
  author       = {Sen Wang and
                  Shao Zeng and
                  Tianjun Gu and
                  Zhizhong Zhang and
                  Ruixin Zhang and
                  Shouhong Ding and
                  Jingyun Zhang and
                  Jun Wang and
                  Xin Tan and
                  Yuan Xie and
                  Lizhuang Ma},
  title        = {From Enhancement to Understanding: Build a Generalized Bridge for
                  Low-light Vision via Semantically Consistent Unsupervised Fine-tuning},
 booktitle    = {ICCV},
  year         = {2025},

}

@inproceedings{Retinexformer23,
  author       = {Yuanhao Cai and
                  Hao Bian and
                  Jing Lin and
                  Haoqian Wang and
                  Radu Timofte and
                  Yulun Zhang},
  title        = {Retinexformer: One-stage Retinex-based Transformer for Low-light Image
                  Enhancement},
  booktitle    = {ICCV},
  year         = {2023},
}

@inproceedings{DiT23,
  author       = {William Peebles and
                  Saining Xie},
  title        = {Scalable Diffusion Models with Transformers},
  booktitle    = {ICCV},
  year         = {2023},
}

@article{ijcv26_uhdpromer,
  author       = {Cong Wang and
                  Jinshan Pan and
                  Liyan Wang and
                  Wei Wang and
                  Yang Yang},
  title        = {Neural Discrimination-Prompted Transformers for Efficient {UHD} Image
                  Restoration and Enhancement},
  journal      = {IJCV},
  volume       = {134},
  number       = {3},
  pages        = {84},
  year         = {2026},
}

@inproceedings{aaai25_pptformer,
  author       = {Cong Wang and
                  Jinshan Pan and
                  Liyan Wang and
                  Wei Wang},
  title        = {Intra and Inter Parser-Prompted Transformers for Effective Image Restoration},
  booktitle    = {AAAI},
  pages        = {7609--7618},
  year         = {2025},
}

@inproceedings{PercepLIE_mm25,
  author       = {Cong Wang and
                  Chengjin Yu and
                  Jie Mu and
                  Wei Wang},
  title        = {PercepLIE: {A} New Path to Perceptual Low-Light Image Enhancement},
  booktitle    = {ACM MM},
  pages        = {6530--6539},
  year         = {2024},
}

@inproceedings{wang2024correlation,
  title={Correlation matching transformation transformers for uhd image restoration},
  author={Wang, Cong and Pan, Jinshan and Wang, Wei and Fu, Gang and Liang, Siyuan and Wang, Mengzhu and Wu, Xiao-Ming and Liu, Jun},
  booktitle={AAAI},
  volume={38},
  number={6},
  pages={5336--5344},
  year={2024}
}

@inproceedings{wang2020joint,
  title={Joint self-attention and scale-aggregation for self-calibrated deraining network},
  author={Wang, Cong and Wu, Yutong and Su, Zhixun and Chen, Junyang},
  booktitle={ACM MM},
  pages={2517--2525},
  year={2020}
}

@inproceedings{msgnn_ijcai24,
  author       = {Cong Wang and
                  Wei Wang and
                  Chengjin Yu and
                  Jie Mu},
  title        = {Explore Internal and External Similarity for Single Image Deraining
                  with Graph Neural Networks},
  booktitle    = {IJCAI},
  pages        = {1371--1379},
  year         = {2024}
}

@article{wang2025ultra,
  title={Ultra-high-definition image restoration: New benchmarks and a dual interaction prior-driven solution},
  author={Wang, Liyan and Wang, Cong and Pan, Jinshan and Liu, Xiaofeng and Zhou, Weixiang and Sun, Xiaoran and Wang, Wei and Su, Zhixun},
  journal={IEEE TCSVT},
  year={2025},
}

@article{uhd_survey,
  author       = {Liyan Wang and
                  Weixiang Zhou and
                  Cong Wang and
                  Kin{-}Man Lam and
                  Zhixun Su and
                  Jinshan Pan},
  title        = {Deep Learning-Driven Ultra-High-Definition Image Restoration: {A}
                  Survey},
  journal      = {CoRR},
  volume       = {abs/2505.16161},
  year         = {2025},
}

@inproceedings{jin2022unsupervised,
  title={Unsupervised night image enhancement: When layer decomposition meets light-effects suppression},
  author={Jin, Yeying and Yang, Wenhan and Tan, Robby T},
  booktitle={ECCV},
  pages={404--421},
  year={2022},
}

@inproceedings{jin2023enhancing,
  title={Enhancing visibility in nighttime haze images using guided apsf and gradient adaptive convolution},
  author={Jin, Yeying and Lin, Beibei and Yan, Wending and Yuan, Yuan and Ye, Wei and Tan, Robby T},
  booktitle={ACM MM},
  pages={2446--2457},
  year={2023}
}
}


\end{document}